\documentclass{article}

    \PassOptionsToPackage{numbers, compress}{natbib}

\usepackage[final]{neurips_2025}

\usepackage[utf8]{inputenc} 
\usepackage[T1]{fontenc}    
\usepackage{url}            
\usepackage{booktabs}       
\usepackage{amsfonts}       
\usepackage{nicefrac}       
\usepackage{microtype}      
\usepackage{xcolor}         
\usepackage{fontawesome5}
\usepackage{booktabs}
\usepackage{tabularx}
\usepackage{subcaption}
\usepackage[table]{xcolor}

\newcommand{\xmark}{{\color{myRed}\ding{55}}}%
\newcommand{\cmark}{{\color{myGreen}\ding{51}}}
\usepackage{xcolor}
\usepackage{pifont} 
\definecolor{myGray}{rgb}{0.5, 0.5, 0.5}
\definecolor{myRed}{rgb}{0.808,0.067,0.149}
\definecolor{myGreen}{rgb}{0.067,0.708,0.149}
\usepackage{multirow}

\usepackage{longtable}
\usepackage[skins]{tcolorbox} 
\tcbuselibrary{breakable}
\usepackage{float}
\newfloat{Prompt}{htbp}{loa}
\usepackage{placeins}
\usepackage{times}
\usepackage{epsfig}
\usepackage{graphicx}
\usepackage{amsmath}
\usepackage{amssymb}
\usepackage{xcolor}
\usepackage{wrapfig}
\definecolor{darkgreen}{rgb}{0.0,0.5,0.0}
\usepackage{booktabs, xcolor, colortbl}
\usepackage{enumitem}

\newcommand{\example}[3]{
\begin{tcolorbox}[colback=white!95!gray, boxrule=0.3pt, sharp corners]
\textbf{Source Narration:} #1\\
\textbf{Target Narration:} #2\\
\textbf{Instruction:} #3
\end{tcolorbox}
}

\usepackage[breaklinks=true,bookmarks=false]{hyperref}

\usepackage[rightcaption]{sidecap}
\usepackage{graphicx}

\begin{document}

\title{From Play to Replay: Composed Video Retrieval for Temporally Fine-Grained Videos
}

\author{
Animesh Gupta$^{1}$ \quad
Jay Parmar$^{1}$ \quad
Ishan Rajendrakumar Dave$^{2}$ \quad
Mubarak Shah$^{1}$ \\
$^{1}$Center for Research in Computer Vision, University of Central Florida \quad
$^{2}$Adobe
}

\maketitle

\begin{abstract}

    Composed Video Retrieval (CoVR) retrieves a target video given a query video and a modification text describing the intended change. Existing CoVR benchmarks emphasize appearance shifts or coarse event changes and therefore do not test the ability to capture subtle, fast-paced temporal differences.    
    We introduce TF-CoVR, the first large-scale benchmark dedicated to temporally fine-grained CoVR. TF-CoVR focuses on gymnastics and diving, and provides 180K triplets drawn from FineGym and FineDiving datasets. 
    Previous CoVR benchmarks, focusing on temporal aspect, link each query to a single target segment taken from the same video, limiting practical usefulness. In TF-CoVR, we instead construct each <query, modification> pair by prompting an LLM with the label differences between clips drawn from different videos; every pair is thus associated with multiple valid target videos (3.9 on average), reflecting real-world tasks such as sports-highlight generation. To model these temporal dynamics, we propose TF-CoVR-Base, a concise two-stage training framework: (i) pre-train a video encoder on fine-grained action classification to obtain temporally discriminative embeddings; (ii) align the composed query with candidate videos using contrastive learning.
    We conduct the first comprehensive study of image, video, and general multimodal embedding (GME) models on temporally fine-grained composed retrieval in both zero-shot and fine-tuning regimes. On TF-CoVR, TF-CoVR-Base improves zero-shot mAP@50 from 5.92 (LanguageBind) to 7.51, and after fine-tuning raises the state-of-the-art from 19.83 to 27.22.
    We have released our dataset and code publicly available at \url{https://github.com/UCF-CRCV/TF-CoVR}.

\end{abstract}

\section{Introduction}
\begin{figure}[t] 
    \centering 
    \includegraphics[width=\textwidth]{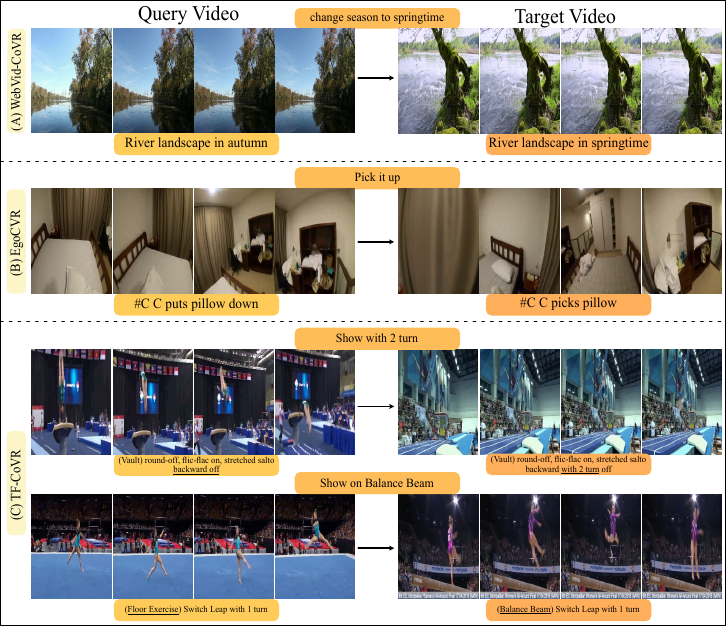} 
    \caption{Comparison of composed-retrieval triplets in WebVid-CoVR, Ego-CVR, and TF-CoVR.
(a) WebVid-CoVR targets appearance changes. (b) Ego-CVR selects the target clip from a different time-stamp of the \emph{same} video, showing a new interaction with the same object. (c) TF-CoVR supports two fine-grained modification types: temporal change- varying sub-actions within the same event (row 3), and event change- the same sub-action performed on different apparatuses (row 4).}
    \label{fig:teaser} 
    \vspace{-2em}
\end{figure}

Recent progress in content‐based image retrieval has evolved into multimodal \emph{composed image retrieval} (CoIR) \cite{yelamarthi2018zero,anwaar2021compositional,liu2016deep}, where a system receives a \emph{query image} and a short \emph{textual modification} and returns the image that satisfies the composition.  
\emph{Composed video retrieval} (CoVR) \cite{ventura2024covr} generalizes this idea, asking for a target video that realizes a user-described transformation of a query clip, for example, ``same river landscape, but in springtime instead of autumn'' (Fig.~\ref{fig:teaser}a) or ``same pillow, but picking up rather than putting down''(Fig.~\ref{fig:teaser}b).  

Existing CoVR benchmarks cover only a limited portion of the composition space. For example, WebVid-CoVR~\cite{ventura2024covr} (Fig.~\ref{fig:teaser}a) is dominated by appearance changes and demands minimal temporal reasoning, while Ego-CVR~\cite{hummel2024egocvr} restricts the query and target to different segments of a single video (Fig.~\ref{fig:teaser}b).  
In practice, many high-value applications depend on \emph{fine-grained} motion differences: surgical monitoring of subtle patient movements \cite{tscholl2020situation}, low-latency AR/VR gesture recognition \cite{xu2021hmd}, and sports analytics where distinguishing a 1.5-turn from a 2-turn somersault drives coaching feedback \cite{hong2021video,naik2022comprehensive}.  
The commercial impact is equally clear: the Olympic Broadcasting Service AI highlight pipeline in Paris 2024 increased viewer engagement 13 times in 14 sports \cite{ioc2024marketing}.  
No public dataset currently evaluates CoVR at this temporal resolution.

To address these limitations, we present \emph{TF-CoVR} (\underline{T}emporally \underline{F}ine-grained \underline{Co}mposed \underline{V}ideo \underline{R}etrieval), a large-scale benchmark for composed retrieval in gymnastics and diving constructed from the temporally annotated FineGym~\cite{shao2020finegym} and FineDiving~\cite{xu2022finediving} datasets.  
Previous work such as Ego-CVR~\cite{hummel2024egocvr} restricts query and target clips to different segments of a \emph{single} video; in practice, however, relevant results often come from distinct videos.  
TF-CoVR instead provides 180K triplets, each containing a query video, a textual modification, and one or more ground-truth target videos.  
We call each \(\langle\)query, modification\(\rangle\) pair a \emph{composed query}.  
The benchmark covers both event-level changes (e.g.\ the same sub-action on different apparatuses) and fine-grained sub-action transitions (e.g.\ varying rotation counts or entry/exit techniques), yielding a setting that reflects real-world temporally fine-grained retrieval far more closely than existing datasets. A thorough comparison with prior datasets is shown in Table~\ref{tab:benchmark_matrix}.

Existing CoVR models, trained on appearance-centric data, usually obtain video representations by simply averaging frame embeddings, thereby discarding temporal structure. Fine-grained retrieval demands video embeddings that preserve these dynamics.  
To this end we introduce a strong baseline, \emph{TF-CoVR-Base}. Unlike recent video-language systems that depend on large-scale descriptive caption rewriting with LLMs, TF-CoVR-Base follows a concise two-stage pipeline. 
\emph{Stage 1} pre-trains a video encoder on fine-grained action classification, producing temporally discriminative embeddings.  
\emph{Stage 2} forms a composed query by concatenating the query-video embedding with the text-modification embedding and aligns it with candidate video embeddings via contrastive learning.

We benchmark TF-CoVR with image-based CoIR baselines, video-based CoVR systems, and general multimodal embedding (GME) models such as E5-V, evaluating every method in both zero-shot and fine-tuned regimes.  
TF-CoVR-Base attains 7.51 mAP@50 in the zero-shot setting, surpassing the best GME model (E5-V, 5.22) and all specialized CoVR methods. Fine-tuning further lifts performance to 27.22 mAP@50, a sizeable gain over the previous state-of-the-art BLIP\(_{\text{CoVR-ECDE}}\) (19.83). These results underscore the need for temporal granularity and motion-aware supervision in CoVR, factors often missing in current benchmarks. TF-CoVR provides the scale to support this and exposes the limitations of appearance-based models.

To summarize, our main contributions are as follows:
\begin{itemize}[leftmargin=*,labelsep=1em,topsep=0pt,itemsep=0pt,parsep=0pt]
\item We introduce \textit{TF-CoVR}, a large-scale benchmark for composed video retrieval centered on sports actions. The dataset comprises 180K training triplets and a test set where each query is associated with an average of 3.9 valid targets, enabling more realistic and challenging evaluation.
\item We propose \textit{TF-CoVR-Base}, a simple yet strong baseline that captures temporally fine-grained visual cues without relying on descriptive, LLM-generated captions.
\item We provide the first comprehensive study of image, video, and GME models on temporally fine-grained composed retrieval under both zero-shot and fine-tuning protocols, where TF-CoVR-Base yields consistent gains across settings.
\end{itemize}

\section{Related Work}
\definecolor{lightblue}{RGB}{230, 245, 255}

\begin{table}[t]
\centering
\caption{Comparison of existing datasets for composed image and video retrieval, highlighting the unique features of TF-CoVR. Datasets are categorized by modality (Type), where \faCamera \hspace{0.3em}indicates image-based and \faVideo \hspace{0.3em} indicates video-based triplets.}
\vspace{0.5em}
\label{tab:benchmark_matrix}
\setlength{\tabcolsep}{5pt}
\resizebox{\linewidth}{!}{
\begin{tabular}{l c c c c c c c}
\toprule
\textbf{Dataset} & \textbf{Type} & \textbf{\#Triplets} & \textbf{Train} & \textbf{Eval} & \textbf{Multi-GT} & \textbf{Eval Metrics} & \textbf{\#Sub-actions} \\
\midrule
CIRR \cite{liu2021image}     & \faCamera   & 36K     & \cmark & \cmark & \xmark      & Recall@K     & \xmark \\
FashionIQ \cite{wu2021fashion}  & \faCamera & 30K     & \cmark & \cmark & \xmark      & Recall@K     & \xmark \\
CC-CoIR  \cite{ventura2024covr}  & \faCamera & 3.3M    & \cmark & \xmark & \xmark      & Recall@K     & \xmark \\
MTCIR  \cite{huynh2025collm}     & \faCamera & 3.4M    & \cmark & \xmark & \xmark      & Recall@K     & \xmark \\
WebVid-CoVR \cite{ventura2024covr} & \faVideo & 1.6M    & \cmark & \cmark & \xmark      & Recall@K     & \xmark \\
EgoCVR  \cite{hummel2024egocvr}     & \faVideo & 2K      & \xmark & \cmark & \xmark      & Recall@K     & \xmark \\
FineCVR~\cite{yue2025learning} & \faVideo & 1M & \cmark & \cmark & \xmark & Recall@K & \xmark \\
CIRCO   \cite{baldrati2023zero}    & \faCamera & 800     & \xmark & \cmark & \cmark      & mAP@K        & \xmark \\
\rowcolor{orange!30}
TF-CoVR (Ours) & \faVideo & 180K    & \cmark & \cmark & \cmark      & mAP@K        & 306 \\
\bottomrule
\end{tabular}
}
\vspace{-1.5em}

\end{table}

\textbf{Video Understanding and Fast-Paced Datasets:} Video understanding \cite{mahmood2024vurf} often involves classifying videos into predefined action categories~\cite{hutchinson2021video, kong2022human, ulhaq2022vision}. These tasks are broadly categorized as coarse- or fine-grained. Coarse-grained datasets like Charades~\cite{sigurdsson2016hollywood} and Breakfast~\cite{kuehne2014language} capture long, structured activities, but lack the temporal resolution and action granularity needed for composed retrieval. In contrast, fine-grained datasets like FineGym~\cite{shao2020finegym} and FineDiving~\cite{xu2022finediving} provide temporally segmented labels for sports actions. They cover high-motion actions where subtle differences (e.g., twists or apparatus) lead to semantic variation, making them suitable for retrieval tasks with fine-grained temporal changes. Yet these datasets remain unexplored in the CoVR setting, leaving a gap in leveraging temporally rich datasets. \textit{TF-CoVR} bridges this gap by introducing a benchmark that explicitly targets temporally grounded retrieval in fast-paced, fine-grained video settings.

\textbf{Composed Image Retrieval:} CoIR retrieves a target image using a query image and a modification text describing the desired change. CoIR models are trained on large-scale triplets of query image, modification text, and target image~\cite{vo2019composing, gu2023compodiff, levy2024data}, which have proven useful for generalizing across open-domain retrieval. CIRR~\cite{liu2021image} provides 36K curated triplets with human-written modification texts for CoIR, but it suffers from false negatives and query mismatches. CIRCO~\cite{baldrati2023zeroshot} improves on this by using COCO~\cite{lin2014microsoft} and supporting multiple valid targets per query. More recently, CoLLM~\cite{huynh2025collm} released MTCIR, a 3.4M triplet dataset with natural captions and diverse visual scenes, addressing the lack of large-scale, non-synthetic data. Despite recent progress, existing CoIR datasets remain inherently image-centric and lack temporal depth, which restricts their applicability to video retrieval tasks requiring fine-grained temporal alignment.

\textbf{Composed Video Retrieval:} WebVid-CoVR~\cite{ventura2024covr} first introduced CoVR as a video extension of CoIR, using query-modification-target triplets sampled from open-domain videos. However, its lack of temporal grounding limits WebVid-CoVR’s effectiveness in retrieving videos based on fine-grained action changes. EgoCVR~\cite{hummel2024egocvr} addressed this by constructing triplets within the same egocentric video to capture temporal cues. FineCVR~\cite{yue2025learning} advanced CoVR by constructing a fine-grained retrieval benchmark using existing video understanding datasets such as ActivityNet~\cite{caba2015activitynet}, ActionGenome~\cite{ji2020action}, HVU~\cite{diba2020large}, and MSR-VTT~\cite{xu2016msr}. Additionally, it introduced a consistency attribute in the modification text to guide retrieval more effectively. While an important step, the source datasets are slow-paced and coarse-grained, limiting their ability to capture subtle action transitions. Despite progress, CoVR benchmarks remain limited, relying mostly on slow-paced or object-centric content and offer only a single target per query, limiting real-world evaluation where multiple valid matches may exist.

\textbf{Multimodal Embedding Models for Composed Retrieval:} Recent advances in MLLMs such as GPT-4o~\cite{hurst2024gpt}, LLaVa~\cite{liu2023visual, liu2024llavanext}, and QwenVL~\cite{wang2024qwen2} have significantly accelerated progress in joint visual-language understanding and reasoning tasks \cite{raza2025responsible, campos2025gaea, Swetha_Xformer_ECCV2024, raza2025humanibench, swetha2023preserving}. VISTA~\cite{zhou2024vista} and MARVEL~\cite{zhou2023marvel} extend image-text retrieval by pairing pre-trained text encoders with enhanced vision encoders to better capture joint semantics. E5-V~\cite{jiang2024e5} and MM-Embed~\cite{lin2024mm} further improve retrieval by using relevance supervision and hard negative mining to mitigate modality collapse. Zhang et al. recently introduced GME~\cite{zhang2024gme}, a retrieval model that demonstrates strong performance on CoIR, particularly in open-domain image-text query settings. However, GME and similar MLLM-based retrievers remain untested in CoVR, especially in fast-paced scenarios requiring fine-grained temporal alignment.

\section{TF-CoVR: Dataset Generation}
\begin{figure}[t] 
    \centering 
    \includegraphics[width=\textwidth]{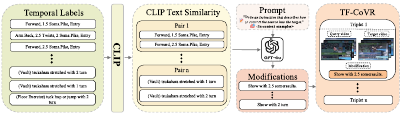} 
    \caption{Overview of our automatic triplet generation pipeline for TF-CoVR. We start with temporally labeled clips from FineGym and FineDiving datasets. Using CLIP-based text embeddings, we compute similarity between temporal labels and form pairs with high semantic similarity. These label pairs are passed to GPT-4o along with in-context examples to generate natural language modifications describing the temporal differences between them. Each generated triplet consists of a query video, a target video, and a modification text capturing fine-grained temporal action changes.}
    \label{fig:dataset-generation} 
    \vspace{-1.5em}
\end{figure}

\textbf{FineGym and FineDiving for Composed Video Retrieval:} Composed video retrieval (CoVR) operates on triplets \((V_q, T_m, V_t)\),  where \(V_q\), \(T_m\), and \(V_t\) denote the query video, modification text, and target video, respectively. Prior works \cite{ventura2024covr, hummel2024egocvr} construct such triplets by comparing captions and selecting pairs that differ by a small textual change, often a single word. This approach, however, relies on the availability of captions, which limits its applicability to datasets without narration. To overcome this, we use FineGym~\cite{shao2020finegym} and FineDiving~\cite{xu2022finediving}, which contain temporally annotated segments but no captions. Instead of captions, we utilize the datasets' fine-grained temporal labels, which describe precise sub-actions. FineGym provides 288 labels over 32,697 clips (avg. 1.7s), from 167 long videos, and FineDiving includes 52 labels across 3,000 clips.

To identify meaningful video pairs, we compute CLIP-based similarity scores between all temporal labels and select those with high semantic similarity \cite{narnaware2025sb}. These pairs are then manually verified and categorized into two types: (1) temporal changes, where the sub-action differs within the same event (e.g., \textit{(Vault) round-off, flic-flac with 0.5 turn on, stretched salto forward with 0.5 turn off} vs. \textit{...with 2 turn off}), and (2) event changes, where the same sub-action occurs in different apparatus contexts (e.g., \textit{(Floor Exercise) switch leap with 1 turn} vs. \textit{(Balance Beam) switch leap with 1 turn}). These examples show that even visually similar actions can have different semantic meanings depending on temporal or contextual cues. We apply this strategy to both FineGym and FineDiving to generate rich, fine-grained video triplets. (See Figure~\ref{fig:teaser} for illustrations.)

\begin{figure}[t] 
    \centering 
    \includegraphics[width=\textwidth]{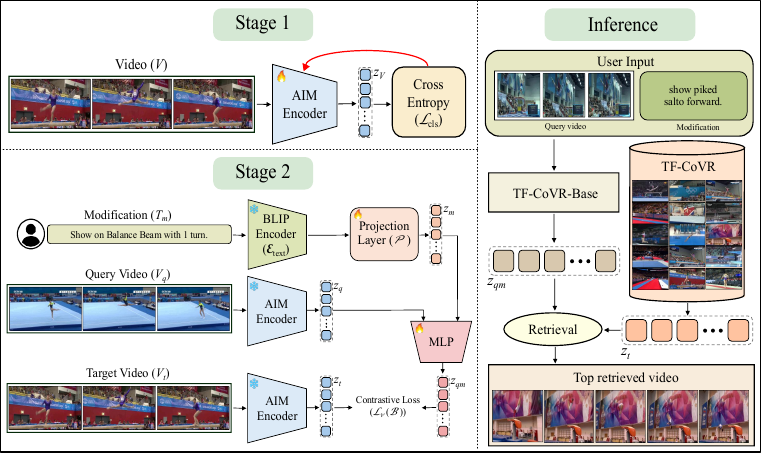} 
    \caption{Overview of TF-CoVR-Base framework. Stage 1 learns temporal video representations via supervised classification using the AIM encoder. In Stage 2, the pretrained AIM and BLIP encoders are frozen, and a projection layer and MLP are trained to align the query-modification pair with the target video using contrastive loss.During inference, the model retrieves relevant videos from TF-CoVR based on a user-provided query video and textual modification.}
    \vspace{-1.5em}
    \label{fig:aim-2stages} 
\end{figure}
\textbf{Modification Instruction and Triplet Generation:} To generate modification texts for TF-CoVR, we start with the fine-grained temporal labels associated with gymnastics and diving segments, such as \textit{Forward, 1.5 Soms.Pike, Entry} or \textit{(Vault) tsukahara stretched with 2 turn}. Using CLIP, we compute pairwise similarity scores between all labels and select those that differ in small but meaningful aspects, representing source and target actions connected by a semantic modification.

Each selected label pair is passed to GPT-4o~\cite{hurst2024gpt} along with a prompt and 15 in-context examples capturing typical sub-action and event-level changes \cite{vayani2024all}. GPT-4o generates concise natural language instructions that describe how to transform the source into the target, e.g., \textit{Show with 2.5 somersaults} or \textit{Show on Balance Beam}. Unlike prior work such as FineCVR~\cite{yue2025learning}, which emphasizes visual consistency, our modifications focus exclusively on temporal changes, making them better suited for real-world use cases like highlight generation where visual similarity is not required.

To form triplets, we split the original long-form videos into training and testing sets to avoid overlap. From these, sub-action clips are extracted and paired with the corresponding modification text. Although individual clips may be reused, each resulting triplet, comprising a query video, a modification text, and a target video, is unique. This process is repeated exhaustively across all labeled segments. Figure~\ref{fig:dataset-generation} illustrates the full pipeline, from label pairing to triplet generation.

\textbf{TF-CoVR Statistics:} TF-CoVR contains 180K training triplets and 473 testing queries, each associated with multiple ground-truth target videos (Table~\ref{tab:benchmark_matrix}). The test set specifically addresses the challenge of evaluating multiple valid retrievals, a limitation in existing CoVR benchmarks. The dataset spans 306 fine-grained sports actions: 259 from FineGym~\cite{shao2020finegym} and 47 from FineDiving~\cite{xu2022finediving}. Clip durations range from 0.03s to 29.00s, with an average of 1.90s.

Modification texts vary from 2 to 19 words (e.g., \textit{“show off”} to \textit{“Change direction to Reverse, reduce to two and a half twists, and show with one and a half somersaults”}), with an average length of 6.11 words. Each test query has an average of 3.94 valid targets, supporting realistic and challenging evaluation under a multi-ground-truth setting. This makes TF-CoVR suited for applications like highlight generation in sports broadcasting, where retrieving diverse sub-action variations is essential.

\section{TF-CoVR-Base: Structured Temporal Learning for CoVR}
\begin{table}[t]
\centering
\caption{Benchmarking results on TF-CoVR using mAP@K for $K \in \{5, 10, 25, 50\}$. We evaluate two groups of models: (1) \textit{Existing CoVR methods trained on WebVid-CoVR and not fine-tuned on TF-CoVR}, and (2) \textit{General Multimodal Embeddings}, tested in a zero-shot setting. Each model is evaluated on query-target pairs consisting of the specified number of sampled frames. ``CA'' denotes the use of cross-attention fusion.}
\vspace{0.5em}
\label{tab:zeroshot_bench}
\setlength{\tabcolsep}{4pt}
\resizebox{\textwidth}{!}{%
\begin{tabular}{
    cc     
    c      
    c      
    c c    
    c c c c
}
\toprule
\multicolumn{2}{c}{\textbf{Modalities}} 
  & \textbf{Model}
  & \textbf{Fusion}
  & \#\textbf{Query}
  & \#\textbf{Target}
  & \multicolumn{4}{c}{\textbf{mAP@K} ($\uparrow$)} \\
\cmidrule(lr){1-2} \cmidrule(lr){7-10}
\textbf{Video} & \textbf{Text} 
  &  &  & \textbf{Frames} & \textbf{Frames}
  & 5 & 10 & 25 & 50 \\
\midrule
\multicolumn{9}{c}{\textit{General Multimodal Embeddings (TF‐CoVR)}} \\
\midrule
\cmark & \cmark
  & GME‐Qwen2‐VL‐2B \cite{zhang2024gme} & MLLM & 1 & 15
  & 2.28 & 2.64 & 3.29 & 3.81 \\ 
\cmark & \cmark
  & MM‐Embed \cite{lin2024mm} & MLLM & 1 & 15
  & 2.39 & 2.81 & 3.61 & 4.14 \\ 
\cmark & \cmark
  & E5‐V \cite{jiang2024e5} & Avg & 1 & 15
  & 3.14 & 3.78 & 4.65 & 5.22 \\ 
\midrule
\multicolumn{9}{c}{\textit{Not fine-tuned on TF-CoVR}} \\
\midrule
\xmark & \cmark
  & BLIP2 & - & - & 15
  & 1.34 & 1.79 & 2.20 & 2.50 \\ 
\cmark & \xmark
& BLIP2 & - & 1 & 15
& 1.74 & 2.20 & 3.06 & 3.62 \\ 
\cmark & \cmark
  & BLIP-CoVR \cite{ventura2024covr} & CA & 1 & 15
  & 2.33 & 2.99 & 3.90 & 4.50 \\  
\cmark & \cmark
  & BLIP$_{\text{CoVR-ECDE}}$ \cite{thawakar2024composed} & CA & 1 & 15
  & 0.78 & 0.88 & 1.16 & 1.37 \\ 
\xmark & \cmark
  & TF‐CVR  \cite{hummel2024egocvr} & - & - & 15
  & 0.56 & 0.76 & 0.99 & 1.24 \\
\cmark & \cmark
  & LanguageBind \cite{zhu2023languagebind} & Avg & 8 & 8
  & 3.43 & 4.37 & 5.26 & 5.92 \\ 
\cmark & \cmark
  & AIM (k400) & Avg & 8 & 8
  & 3.75 & 4.37 & 5.47 & 6.12 \\ 
\cmark & \cmark
  & AIM (k400) & Avg & 16 & 16
  & 4.23 & 5.14 & 6.37 & 7.13 \\ 
\cmark & \cmark
  & AIM (k400) & Avg & 32 & 32
  & 4.22 & 5.15 & 6.50 & 7.30 \\ 
\cmark & \cmark
  & AIM (diving48) & Avg & 32 & 32
  & \textbf{4.81} & \textbf{5.78} & \textbf{6.82} & \textbf{7.51} \\ 
\bottomrule
\end{tabular}
}
\vspace{-1.5em}
\end{table}

\textbf{Method Overview:} 
In the composed video retrieval (CoVR) task, the goal is to retrieve a target video \( V_t \) given a query video \( V_q \) and a textual modification \( T_m \) that describes the intended transformation. This requires learning a cross-modal relationship between visual and textual inputs that captures how the target differs from the query. While prior methods have shown promise on general video datasets, TF-CoVR becomes significantly more challenging in fine-grained, fast-paced domains such as gymnastics and diving, where subtle temporal action differences are critical. Existing approaches often overlook these dynamics, motivating the need for a more temporally grounded framework.

\textbf{Two-Stage CoVR Approach:} We propose a two-stage training framework, TF-CoVR-Base, for composed video retrieval in fine-grained, fast-paced domains such as gymnastics and diving. TF-CoVR-Base is designed to explicitly capture the temporal structure in videos and align it with textual modifications for accurate retrieval. Unlike prior approaches that rely on average-pooled frame features from image-level encoders, TF-CoVR-Base decouples temporal representation learning from the retrieval task. It first learns temporally rich video embeddings through supervised action classification, and then uses these embeddings in a contrastive retrieval setup. We describe each stage of the framework below.

\textbf{Stage One: Temporal Pretraining via Video Classification:}
In the first stage, we aim to learn temporally rich video representations from TF-CoVR. To this end, we employ the AIM encoder~\cite{yang2023aim}, which is specifically designed to capture temporal dependencies by integrating temporal adapters into a CLIP-based backbone.

We pretrain the AIM encoder on a supervised video classification task using all videos from the triplets in the training set. Let \( V = \{f_1, f_2, \ldots, f_f\} \) denote a video clip with \textit{f} frames. The AIM encoder processes each frame and produces a sequence-level embedding:
\[
z_V = \text{AIM}(V).
\]

The classification logits \( z_V\)  are passed through a softmax function to produce a probability distribution over classes:
\[
\hat{p}_V^{(i)} = \text{Softmax}(z_V^{(i)}).
\]
Each video \( V \) is annotated with a ground-truth label \( y_V \), and the model is optimized using the standard cross-entropy loss:
\[
\mathcal{L}_{\text{cls}} = -\sum_{i=1}^{C} y_V^{(i)} \log \hat{p}_V^{(i)}.
\]

where \( C = 306 \) is the total number of fine-grained action classes in the TF-CoVR dataset.

\textbf{Stage Two: Contrastive Training for Retrieval:} In the second stage of TF-CoVR-Base, we train a contrastive model to align the composed query representations with the target video representations. As illustrated in Figure~\ref{fig:aim-2stages}, each training sample is structured as a triplet \( (V_q, T_m, V_t) \), where \( V_q \) is the \textbf{query video} consisting of \( N \) frames, \( T_m \) is the \textbf{modification text} with \( L \) tokens, and \( V_t \) is the \textbf{target video} comprising \( M \) frames.

We use our pretrained and frozen AIM encoder from stage 1 to extract temporally rich embeddings for the query and target videos:
\[
z_q = \text{AIM}(V_q), \quad z_t = \text{AIM}(V_t).
\]

The modification text \(T_m\) is encoded using the BLIP2 text encoder \( \mathcal{E}_{\text{text}} \), followed by a learnable projection layer \( \mathcal{P} \) that maps the text embedding into a shared embedding space. This step ensures the textual features are adapted and aligned with the video modality for the CoVR task:
\[
z_m = \mathcal{P}(\mathcal{E}_{\text{text}}(T_m)).
\]

We then fuse the query video embedding \(z_q\) and the projected text embedding \(z_m\) using a multi-layer perceptron (MLP), producing the composed query representations:
\[
z_{qm} = \text{MLP}(z_q, z_m).
\]

To compare the composed query embeddings with the target video embeddings, both \( z_{qm} \) and \( z_t \) are projected into a shared embedding space and normalized to unit vectors. Their relationship is then measured using cosine, computed as:
\[
S_{i,j} = \frac{z_{qm}^{(i)} \cdot z_t^{(j)}}{\|z_{qm}^{(i)}\| \, \|z_t^{(j)}\|}.
\]

To ensure numerical stability and regulate the scale of similarity scores, cosine similarity is adjusted using a temperature parameter:
\[
\text{sim}(z_{qm}^{(i)}, z_t^{(j)}) = \frac{S_{i,j}}{\tau}.
\]
where \( \tau \in \mathbb{R}_{>0} \) is the temperature parameter. We then define a scaled similarity matrix $\tilde{S}$ using a concentration parameter $\beta \ge 0$:


\begin{align*}
    \tilde{S}_{i,j} = \beta \cdot S_{i,j}.
\end{align*}

The weight assigned to each negative sample in the loss is computed using a softmax-like reweighting scheme, with diagonal entries (positive pairs) scaled by a hyperparameter $\alpha \in (0, 1]$:

\begin{equation*}
\begin{minipage}{0.48\textwidth}
\begin{align*}
    w^{i \rightarrow t}_{i,j} = 
    \begin{cases}
        \alpha, & \text{if } i = j \\
        \displaystyle \frac{(n - 1) \cdot \exp(\tilde{S}_{i,j})}{\sum\limits_{k \ne i} \exp(\tilde{S}_{i,k})}, & \text{otherwise}
    \end{cases}
\end{align*}
\end{minipage}
\hfill
\begin{minipage}{0.48\textwidth}
\begin{align*}
    w^{t \rightarrow i}_{j,i} = 
    \begin{cases}
        \alpha, & \text{if } j = i \\
        \displaystyle \frac{(n - 1) \cdot \exp(\tilde{S}_{j,i})}{\sum\limits_{k \ne i} \exp(\tilde{S}_{k,i})}, & \text{otherwise}
    \end{cases}
\end{align*}
\end{minipage}
\end{equation*}

Finally, the HN-NCE loss~\cite{radenovic2023filtering} is defined as followed, which emphasizes hard negatives by assigning greater weights to semantically similar but incorrect targets. Given a batch \( \mathcal{B} \) of triplets \( (q_i, m_i, t_i) \), the loss is defined as:

\begin{align*}
    \mathcal{L}_v(\mathcal{B}) = \frac{1}{n} \sum_{i=1}^n \left[
    \log \left( \sum_{j=1}^n \exp(S_{i,j}) \cdot w^{i \rightarrow t}_{i,j} \right) +
    \log \left( \sum_{j=1}^n \exp(S_{j,i}) \cdot w^{t \rightarrow i}_{j,i} \right)
    - 2 S_{i,i}
    \right].
\end{align*}

Here, \( S_{i,j} \) is the cosine similarity between the composed query \( z_{qm}^{(i)} \) and the target video \( z_t^{(j)} \), \( \alpha \) is a scalar constant (set to 1), \( \tau \) is a temperature hyperparameter (set to 0.07). In our experiments, we set $\alpha = 1$ and $\beta = 0$, reducing the formulation to the standard InfoNCE \cite{oord2018representation} loss.

\section{Discussion}
\begin{table}[t]
\centering
\caption{Evaluation of models fine-tuned on TF-CoVR using mAP@K for $K \in \{5, 10, 25, 50\}$. We report the performance of various fusion strategies and model architectures trained on TF-CoVR. Fusion methods include MLP and cross-attention (CA). Each model is evaluated using a fixed number of sampled frames from both query and target videos. Fine-tuning on TF-CoVR leads to significant improvements across all models. The results for TF-CoVR-Base (Stage-2 only) reflect the model's performance without Stage-1 temporal pretraining.}
\vspace{0.5em}
\label{tab:table3}
\setlength{\tabcolsep}{3pt}
\resizebox{\textwidth}{!}{%
\begin{tabular}{
    cc     
    c      
    c      
    c c    
    c c c c
}
\toprule
\multicolumn{2}{c}{\textbf{Modalities}} 
  & \textbf{Model}
  & \textbf{Fusion}
  & \#\textbf{Query}
  & \#\textbf{Target}
  & \multicolumn{4}{c}{\textbf{mAP@K ($\uparrow$)}} \\
\cmidrule(lr){1-2} \cmidrule(lr){7-10}
\textbf{Video} & \textbf{Text} 
  &  &  & \textbf{Frames} & \textbf{Frames}
  & 5 & 10 & 25 & 50 \\
\midrule
\multicolumn{9}{c}{\textit{Fine‐tuned on TF-CoVR}} \\
\midrule
\xmark & \cmark
  & BLIP2 & - & - & 15
  & 10.69 & 13.02 & 15.35 & 16.41 \\ 
\cmark & \xmark
  & BLIP2 & - & 1 & 15
  & 4.86 & 6.49 & 8.92 & 10.06 \\ 
\cmark & \cmark
  & CLIP & MLP & 1 & 15
  & 7.01 & 8.35 & 10.22 & 11.38 \\ 
\cmark & \cmark
  & BLIP2 & MLP & 1 & 15
  & 10.86 & 13.20 & 15.38 & 16.31 \\
\cmark & \cmark
  & CLIP & MLP & 15 & 15
  & 6.40 & 7.46 & 9.21 & 10.40 \\ 
\cmark & \cmark
  & BLIP2 & MLP & 15 & 15
  & 11.64 & 14.81 & 16.74 & 17.55 \\
\cmark & \cmark
  & BLIP-CoVR & CA \cite{ventura2024covr} & 1 & 15
  & 11.07 & 13.94 & 16.07 & 16.88 \\
\cmark & \cmark
  & BLIP$_{\text{CoVR-ECDE}}$ & CA \cite{thawakar2024composed} & 1 & 15
  & 13.03 & 15.90 & 18.62 & 19.83 \\ 
\cmark & \cmark
  & TF-CoVR-Base (Stage-2 only) & MLP & 8 & 8
  & 15.08 & 18.70 & 21.78 & 22.61 \\ 
\rowcolor{orange!30}
\cmark & \cmark
  & TF-CoVR-Base (Ours) & MLP & 12 & 12
  & \textbf{21.85} & \textbf{24.23} & \textbf{26.47} & \textbf{27.22} \\ 
\bottomrule
\end{tabular}
}

\end{table}
\begin{wraptable}{r}{0.6\textwidth} 
\centering
\vspace{-1em}
\caption{Performance of GME models on existing CoIR benchmarks. We report mAP@5 and Recall@10 on FashionIQ, CIRR, and CIRCO using official evaluation protocols. Values are directly taken from the original papers.}
\label{tab:finetune_gme}
\setlength{\tabcolsep}{3pt}
\resizebox{0.6\textwidth}{!}{%
\begin{tabular}{c c c c c}
\toprule
\textbf{Model} & \textbf{Metric} & \textbf{FQ} & \textbf{CIRR} & \textbf{CIRCO} \\
\midrule
E5-V~\cite{jiang2024e5} & Recall@10 & 3.73 & 13.19 & - \\
GME-2B~\cite{zhang2024gme} & Recall@10 & 26.34 & 47.70 & - \\
MM-Embed~\cite{lin2024mm} & Recall@10 & 25.7 & 50.0 & - \\
E5-V~\cite{jiang2024e5} & mAP@5 & - & - & 19.1 \\
MM-Embed~\cite{lin2024mm} & mAP@5 & - & - & 32.3 \\
\bottomrule
\end{tabular}
}
\vspace{-2em}
\end{wraptable}

\textbf{Evaluation Metric:} To effectively evaluate retrieval performance in the presence of multiple ground-truth target videos, we adopt the mean Average Precision at \(K\) (mAP@K) metric, as proposed in CIRCO~\cite{baldrati2023zero}. The mAP@K metric measures whether the correct target videos are retrieved and considers the ranks at which they appear in the retrieval results.

Here, \(K\) denotes the number of top-ranked results considered for evaluation. For example, mAP@5 measures precision based on the top 5 retrieved videos, capturing how well the model retrieves relevant targets early in the ranked list. A higher \(K\) allows evaluation of broader retrieval quality, while a lower \(K\) emphasizes top-ranking precision.

\textbf{Specialized vs. Generalized Multimodal Models for CoVR:} We compare specialized models trained specifically for composed video retrieval, such as those trained on WebVid-CoVR~\cite{ventura2024covr}, with Generalized Multimodal Embedding (GME) models that have not seen CoVR data. Among the specialized baselines, we include two image-based encoders (CLIP and BLIP) and one video-based encoder (LanguageBind) to cover different modality types and fusion mechanisms. As shown in Table~\ref{tab:zeroshot_bench}, our evaluation reveals that GME models consistently outperform most specialized CoVR methods in the zero-shot setting. For example, E5-V~\cite{jiang2024e5} achieves 5.22 mAP@50, outperforming BLIP-CoVR (4.50) and BLIP$_{\text{CoVR-ECDE}}$ (1.37), and closely matching LanguageBind (5.92). Other GME variants like MM-Embed and GME-Qwen2-VL-2B also show promising results. In contrast, TF-CVR~\cite{hummel2024egocvr} performs worst among all tested models, with only 1.24 mAP@50, underscoring its limitations in handling fine-grained action variations.

This performance gap is partly due to TF-CVR’s reliance on a captioning model to describe the query video. We replaced the original Lavila~\cite{zhao2023learning} with Video-XL~\cite{shu2024video}, which provides better captions for structured sports content. However, even Video-XL fails to capture subtle temporal cues like twist counts or somersaults, critical for accurate retrieval, causing TF-CVR to struggle with temporally precise matches. In contrast, GME models benefit from large-scale multimodal training involving text, images, and combinations thereof, allowing them to generalize well to CoVR without task-specific fine-tuning. We expect their performance to improve further with fine-tuning on TF-CoVR, though we leave this exploration to future work. See supplementary material for a comparison of Lavila-generated captions.

\begin{figure}[t] 
    \centering 
    \includegraphics[width=\textwidth]{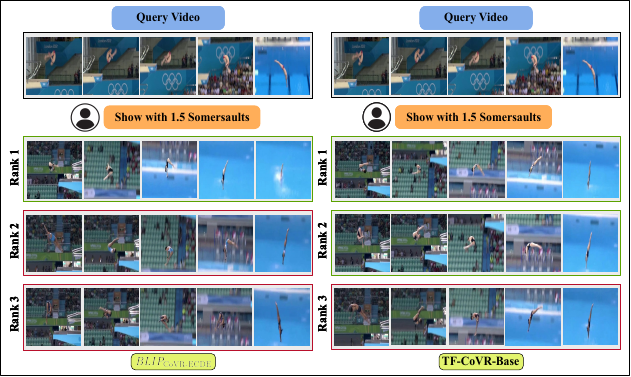} 
    \caption{Qualitative results for the composed video retrieval task using our two-stage TF-CoVR-Base model. Each column showcases a query video (top), the corresponding modification instruction (middle), and the top-3 retrieved target videos (ranks 1–3) based on model predictions. TF-CoVR-Base effectively captures subtle temporal variations and retrieves the correct target video at higher ranks. In contrast, the baseline method BLIP\textsubscript{CoVR-ECDE} often fails to identify the correct action class or resolve fine-grained temporal differences, as indicated by the errors highlighted in \textcolor{red}{red}. }
    \label{fig:aim-qualitative} 
    \vspace{-2em}
\end{figure}

\textbf{Evaluating TF-CoVR-Base Against Existing Methods:} We compare our proposed two-stage TF-CoVR-Base framework with all existing CoVR baselines in Table~\ref{tab:table3}. Our full model achieves 27.22 mAP@50, significantly outperforming the strongest prior method, BLIP$_{\text{CoVR-ECDE}}$ (19.83). Even our Stage-2-only variant (trained without temporal pretraining) outperforms all existing methods with 22.61 mAP@50, highlighting the strength of our contrastive fusion strategy. Unlike BLIP$_{\text{CoVR-ECDE}}$, our model does not rely on detailed textual descriptions of the query video and instead learns temporal structure directly from the visual input. This makes it especially effective in structured, fast-paced sports videos, where subtle action distinctions, such as change in twist count or apparatus, are visually grounded. Across all K values, TF-CoVR-Base shows consistent improvements of 4-6 mAP points.

\begin{wraptable}{r}{0.62\textwidth} 
\centering
\vspace{-1em}
\caption{Performance comparison between the HN-NCE and InfoNCE loss by varying the HN-weighting.}
\label{tab:table5}
\setlength{\tabcolsep}{3pt}
\resizebox{0.6\textwidth}{!}{%
\begin{tabular}{c c c c c}
\toprule
\textbf{HN-Weighting} & \textbf{mAP@5} & \textbf{mAP@10} & \textbf{mAP@25} & \textbf{mAP@50} \\
\midrule
0.7 & 20.40 & 22.46 & 24.63 & 25.37 \\
0.5 & 21.02 & 22.89 & 25.21 & 25.91 \\
0.3 & 20.86 & 23.35 & 25.44 & 26.16 \\
0.0 & 21.85 & 24.23 & 26.47 & 27.22 \\
\bottomrule
\end{tabular}
}
\vspace{-1em}
\end{wraptable}

\textbf{Impact of Hard-Negative Weighting on TF-CoVR:} We further investigate the impact of hard-negative (HN) weighting in the HN-NCE loss function. Specifically, we compare different weighting values, including the baseline setting of 0, which reduces the loss to the standard InfoNCE \cite{oord2018representation} formulation. Our results show that InfoNCE (HN-weighting = 0) consistently outperforms the HN-NCE variants with positive weighting values. While HN-NCE is designed to emphasize hard negatives by assigning them higher weights, this approach can introduce optimization noise, particularly in fine-grained settings where many negative samples are visually similar to the positives. In such scenarios, treating all negatives equally, as in InfoNCE, appears to provide more stable training and better discrimination based on subtle visual cues. As shown in Table~\ref{tab:table5}, reducing the HN-weighting from 0.7 to 0.0 results in a performance gain from 25.37 to 27.22 mAP, an increase of over 1.8 mAP points.

\textbf{Qualitative Analysis:} Figure~\ref{fig:aim-qualitative} illustrates the effectiveness of our method using qualitative examples. The retrieved target videos accurately reflect the action modifications described in the input text. Correctly retrieved clips are outlined in \textcolor{green}{green}, and incorrect ones in \textcolor{red}{red}. Interestingly, even incorrect predictions are often semantically close to the intended target, revealing the fine-grained difficulty of TF-CoVR. For example, in the third column of Figure~\ref{fig:aim-qualitative}, the query video includes a turning motion, while the modification requests a \textit{“no turn”} variation. Our method correctly retrieves \textit{“no turn”} actions at top ranks, but at rank 3, retrieves a “split jump” video, visually similar but semantically different. We highlight this with a red overlay to emphasize the subtle distinction in motion, showing the value of TF-CoVR for evaluating fine-grained temporal reasoning.

\textbf{Domain-Specific Pretraining for Temporal Reasoning:} Although TF-CoVR-Base is designed to be \textit{domain agnostic}, its current training leverages domain-specific datasets to better capture the fine-grained and structured nature of different activity domains, such as surgery or daily tasks. Domain-specific pretraining proves beneficial for learning distinct temporal patterns and visual cues inherent to each domain. For example, in a surgical setting, a query video may depict a sequence such as \textit{``insert needle at a 30-degree angle, advance 2 cm, then begin the suture loop with the right hand,''} while the corresponding target video modifies this to \textit{``insert needle at a 45-degree angle, advance 3 cm, then begin the suture loop with the right hand.''} The modification text, \textit{``change needle insertion angle to 45 degrees and advance by 3 cm instead of 2 cm,''} captures subtle changes in motion angle and depth. Accurately modeling such fine-grained temporal variations necessitates temporally discriminative features, which are challenging to learn without domain-specific pretraining. This positions TF-CoVR-Base to provide a strong foundation for exploring more generalizable temporal reasoning methods across diverse and less-structured video domains.

\section{Limitations and Conclusion}
\textbf{Limitations.} TF-CoVR offers a new perspective on composed video retrieval by focusing on retrieving videos that reflect subtle action changes, guided by a modification text. While it adds valuable depth to the field, the dataset has some limitations. One limitation is that it requires expert effort to temporally annotate videos such as from FineGym and FineDiving, which is currently lacking in the video-understanding community, and such annotation is expensive to scale up. This reflects the trade-off between expert-driven annotations and scalability. Regarding the TF-CoVR-Base, it is currently two-stage, which may not provide a fully end-to-end solution; a better approach could be a single-stage model that simultaneously learns temporally rich video representations and aligns them with the modification text. 

\textbf{Conclusion.} In this work, we introduced TF-CoVR, a large-scale dataset comprising 180K unique triplets centered on fine-grained sports actions, spanning 306 diverse sub-actions from gymnastics and diving videos. TF-CoVR brings a new dimension to the CoVR task by emphasizing subtle temporal action changes in fast-paced, structured video domains. Unlike existing CoVR datasets, it supports multiple ground-truth target videos per query, addressing a critical limitation in current benchmarks and enabling more realistic and flexible evaluation. In addition, we propose a two-stage training framework that explicitly models temporal dynamics through supervised pre-training. Our method significantly outperforms existing approaches on TF-CoVR. Furthermore, we conducted a comprehensive benchmarking of both existing CoVR methods and General Multimodal Embedding (GME) models, marking the first systematic evaluation of GME performance in the CoVR setting. We envision TF-CoVR serving as a valuable resource for real-world applications such as sports highlight generation, where retrieving nuanced sub-action variations is essential for generating engaging and contextually rich video content.

\newpage


{\small
\bibliographystyle{ieee_fullname}
\bibliography{egbib}
}

\newpage

\appendix

\begin{center}
  {\LARGE \bf Supplementary Material \par}
  \vspace{1em}
\end{center}

\renewcommand{\thefigure}{\thesection\arabic{figure}}
\renewcommand{\thetable}{\thesection\arabic{table}}

\setcounter{figure}{0}
\setcounter{table}{0}

\section{TF-CoVR Statistics and Modification Lexicon}

\begin{figure}[h] 
    \centering 
    \includegraphics[width=\textwidth]{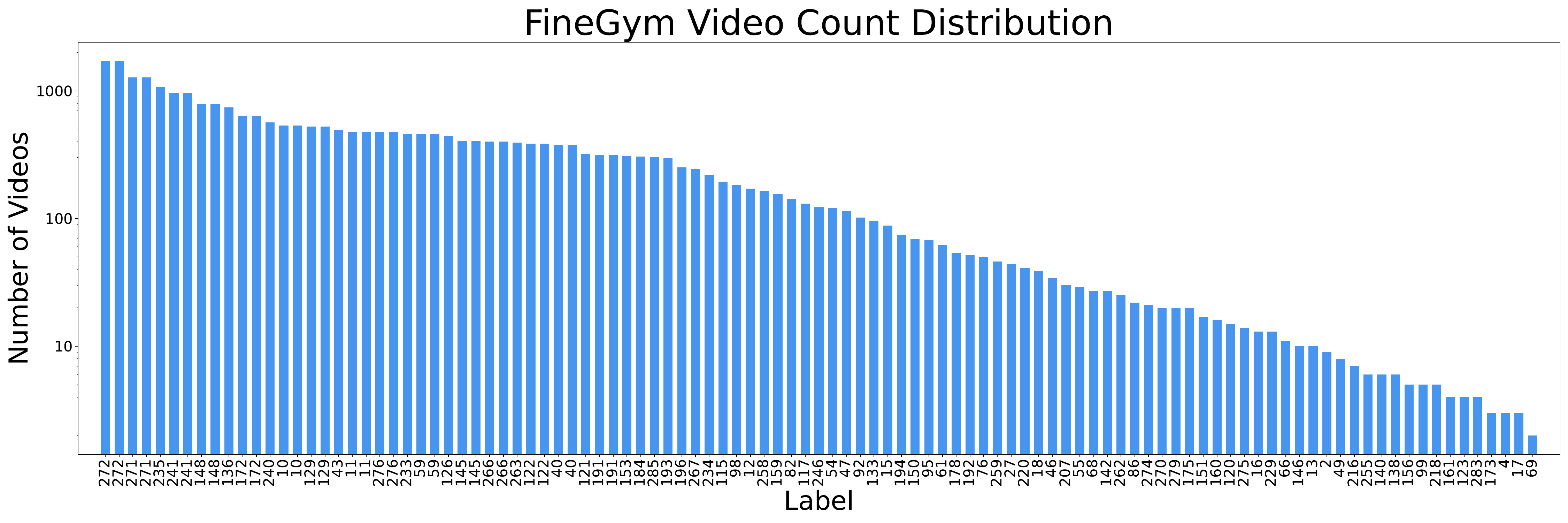} 
    \caption{Label-wise video count distribution in the \textit{FineGym} subset of \textit{TF-CoVR}. A logarithmic scale is used on the y-axis to highlight the steep drop in video counts per label due to the smaller dataset size. Note that only a subset of all labels is shown for clarity.}
    \label{fig:finegym_vc} 
\end{figure}

\begin{figure}[h] 
    \centering 
    \includegraphics[width=\textwidth]{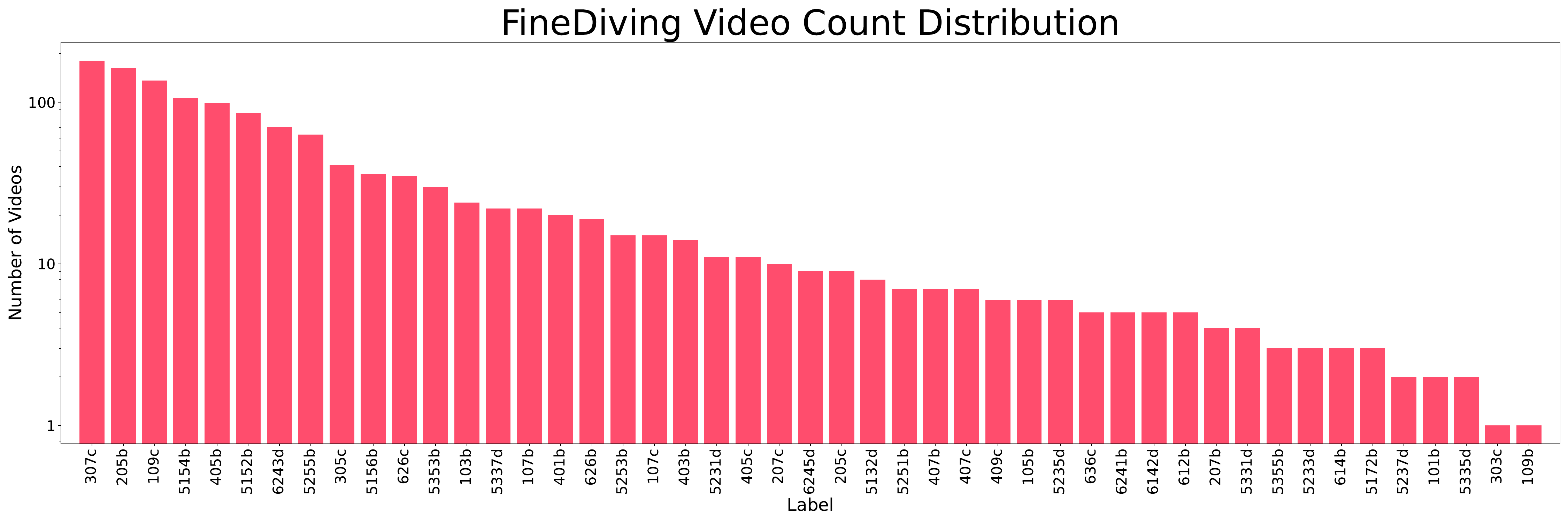} 
    \caption{Label-wise video count distribution in the \textit{FineDiving} subset of \textit{TF-CoVR}. The y-axis is plotted on a logarithmic scale to highlight the steep drop in video counts per label due to the smaller dataset size, while still preserving label diversity.}
    \label{fig:finediving_vc} 
\end{figure}

\paragraph{TF-CoVR Statistics}
We present detailed statistics on the distribution of video counts per label in \textit{TF-CoVR}, which comprises a diverse set of 306 annotated sub-actions. Figures~\ref{fig:finegym_vc} and~\ref{fig:finediving_vc} show the label-wise video distribution for the \textit{FineGym}~\cite{shao2020finegym} and \textit{FineDiving}~\cite{xu2022finediving} subsets of \textit{TF-CoVR}, respectively. Both distributions are plotted on a logarithmic scale to emphasize the long-tailed nature of label frequencies. In \textit{FineGym}, many labels have several hundred to over a thousand associated videos, with a gradual decline across the distribution. This results in broad coverage of fine-grained sub-actions. By contrast, \textit{FineDiving} exhibits a steeper drop in video count per label, primarily due to its smaller dataset size. Nevertheless, a substantial number of labels still contain more than 30 samples, preserving enough diversity to support \textit{temporal fine-grained composed video retrieval.} \textit{TF-CoVR} thus serves as a strong benchmark for learning and evaluating fine-grained temporal reasoning in the composed video retrieval task. 


\vspace{-0.5em}

\vspace{-1em}

\begin{wrapfigure}{r}{0.62\textwidth}
    \centering
    \vspace{-2em}
    \includegraphics[width=0.6\textwidth, trim=125 30 125 30, clip]{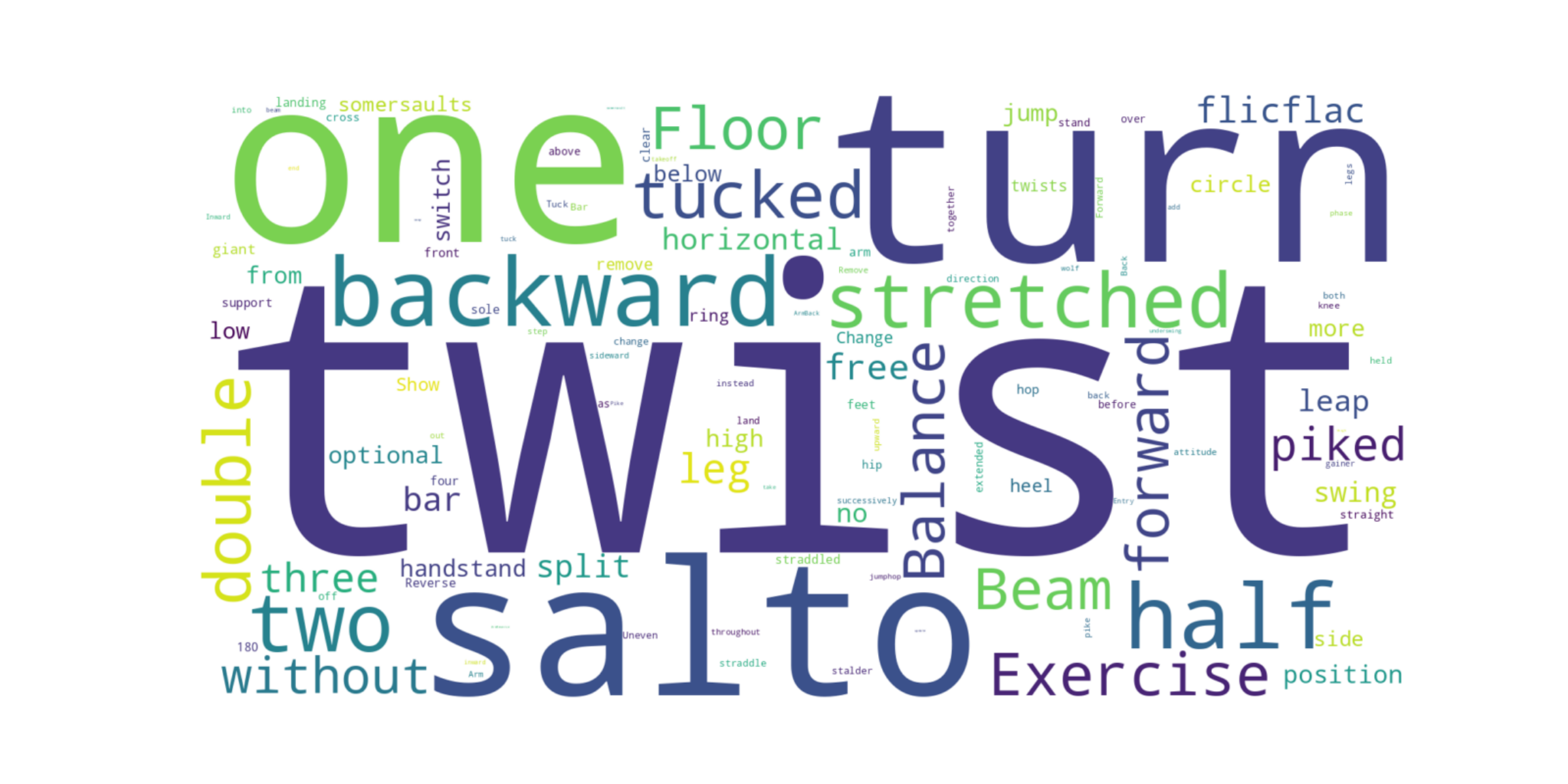}
    \caption{Word cloud visualization of the most frequent action-related terms in \textit{TF-CoVR} modification texts. Larger words indicate higher frequency and reflect the dataset’s fine-grained, motion-centric nature, with terms like \textit{twist}, \textit{turn}, \textit{salto}, and apparatus names such as \textit{Beam} and \textit{Floor} highlighting contextual diversity across domains.}
    \label{fig:wordcloud}
    \vspace{-4em}
\end{wrapfigure}

\paragraph{Modification Lexicon}
Figure~\ref{fig:wordcloud} presents a word cloud visualization of the most frequently occurring terms in the modification texts of \textit{TF-CoVR}. Prominent terms such as \textit{twist}, \textit{turn}, \textit{salto}, \textit{backward}, \textit{tucked}, \textit{stretched}, and \textit{piked} highlight the fine-grained, motion-centric nature of these modifications. These terms encapsulate key action semantics related to orientation, body posture, and movement complexity, covering aspects such as the number of twists or turns, in-air body position, and directional shifts like forward or backward. The presence of apparatus-specific terms such as \textit{Beam}, \textit{Floor}, and \textit{Exercise} further underscores the diversity of event contexts represented in the dataset. This rich and structured lexicon enables \textit{TF-CoVR} to support nuanced temporal modifications, distinguishing it from existing datasets that often rely on coarser or less temporally dynamic instructions.

In this appendix, we provide more details, experimental results, qualitative visualization of our new \textit{TF-CoVR} dataset and our two-stage \textit{TF-CoVR-Base} method.

\section{TF-CoVR: Modification Text Generation}

To support \textit{TF-CoVR} modification generation, we craft domain-adapted prompting strategies for GPT-4o~\cite{hurst2024gpt}, addressing the unique structure of gymnastics and diving videos. Given the structural differences between \textit{FineGym}~\cite{shao2020finegym} and \textit{FineDiving}~\cite{xu2022finediving}, we developed separate prompts for each domain. \textit{FineGym}, with its substantially larger set of annotated sub-actions, was provided with 20 in-context examples to better capture the diversity and complexity of its routines. In contrast, we used 5 in-context examples for \textit{FineDiving}, reflecting its smaller label set and more compact structure.

\paragraph{Prompt and In-Context Examples}
To support accurate modification generation for \textit{TF-CoVR}, we designed prompt templates and in-context examples that align with the linguistic and structural characteristics of the gymnastics and diving domains.

\begin{tcolorbox}[title=\textbf{Modification Generation Prompt for FineDiving}, colback=gray!5, colframe=black!30, coltitle=black, fonttitle=\bfseries, sharp corners, boxrule=0.5pt, arc=3pt]
You are an expert in designing tasks that require understanding the transformation between two description, specifically for video descriptions. Your goal is to ensure that the instructions you provide are concise, accurate, and focused on the necessary modifications between the source and target description.

\textbf{Instructions:}
\begin{enumerate}[leftmargin=1.5em]
    \item Analyze the given source and target description.
    \item Identify the changes between the source and target description.
    \item Write an instruction that describes only the transformation required to achieve the target description from the source.
    \item Ensure the instruction is as short as possible, focusing on actions. Mention objects only when absolutely necessary.
    \item Do not describe objects or actions common to both descriptions. Use pronouns when appropriate.
    \item Your response should focus only on the transformation, without extraneous details or repetitions.
\end{enumerate}

\textbf{Remember:}
\begin{itemize}[leftmargin=1.5em]
    \item Keep the instruction concise and focus only on the transformation required.
    \item Avoid redundant details or describing elements unchanged between source and target descriptions.
\end{itemize}

\end{tcolorbox}

\newpage
\textbf{In-Context Examples:}
\begin{tcolorbox}[colback=white!95!gray, boxrule=0.3pt, sharp corners]
\textbf{Source Description:} Inward, 3.5 Soms.Tuck, Entry\\
\textbf{Target Description:} Inward, 4.5 Soms.Tuck, Entry\\
\textbf{Modification text:} Show with 4.5 somersaults Tuck.
\end{tcolorbox}

\begin{tcolorbox}[colback=white!95!gray, boxrule=0.3pt, sharp corners]
\textbf{Source Description:} Inward, 3.5 Soms.Tuck, Entry\\
\textbf{Target Description:} Inward, 2.5 Soms.Tuck, Entry\\
\textbf{Modification text:} Show with 2.5 somersaults Tuck.
\end{tcolorbox}

\begin{tcolorbox}[colback=white!95!gray, boxrule=0.3pt, sharp corners]
\textbf{Source Description:} Back, 1.5 Twists, 2.5 Soms.Pike, Entry\\
\textbf{Target Description:} Back, 2.5 Twists, 1.5 Soms.Pike, Entry\\
\textbf{Modification text:} Show with 2.5 twists and 1.5 somersaults.
\end{tcolorbox}

\begin{tcolorbox}[colback=white!95!gray, boxrule=0.3pt, sharp corners]
\textbf{Source Description:} Forward, 3.5 Soms.Pike, Entry\\
\textbf{Target Description:} Forward, 1.5 Soms.Pike, Entry\\
\textbf{Modification text:} Show with 1.5 somersaults.
\end{tcolorbox}

\begin{tcolorbox}[colback=white!95!gray, boxrule=0.3pt, sharp corners]
\textbf{Source Description:} Arm.Back, 2.5 Twists, 2 Soms.Pike, Entry\\
\textbf{Target Description:} Arm.Back, 1.5 Twists, 2 Soms.Pike, Entry\\
\textbf{Modification text:} Show with 1.5 twists.
\end{tcolorbox}

\vspace{1em}

\begin{tcolorbox}[title=\textbf{Modification Generation Prompt for FineGym}, colback=gray!5, colframe=black!30, coltitle=black, fonttitle=\bfseries, sharp corners, boxrule=0.5pt, arc=3pt]
You are an expert in designing tasks that require understanding the transformation between two description, specifically for video descriptions. Your goal is to ensure that the instructions you provide are concise, accurate, and focused on the necessary modifications between the source and target description.

\vspace{1em}

\textbf{Instructions:}
\begin{enumerate}[leftmargin=1.5em]
    \item Analyze the given source and target description.
    \item Identify the changes between the source and target description.
    \item Write an instruction that describes only the transformation required to achieve the target description from the source.
    \item Ensure the instruction is as short as possible, focusing on actions. Mention objects only when absolutely necessary.
    \item Do not describe objects or actions common to both descriptions. Use pronouns when appropriate.
    \item Your response should focus only on the transformation, without extraneous details or repetitions.
\end{enumerate}

\textbf{Remember:}
\begin{itemize}[leftmargin=1.5em]
    \item Keep the instruction concise and focus only on the transformation required.
    \item Avoid redundant details or describing elements unchanged between source and target descriptions.
\end{itemize}

\end{tcolorbox}

\vspace{1em}

\textbf{In-Context Examples:}

\example{(VT) round-off, flic-flac with 0.5 turn on, stretched salto forward with 1.5 turn off.}
        {(VT) round-off, flic-flac with 0.5 turn on, stretched salto forward with 0.5 turn off.}
        {show with 0.5 turn.}

\example{(VT) round-off, flic-flac with 0.5 turn on, stretched salto forward with 1.5 turn off.}
        {(VT) round-off, flic-flac with 0.5 turn on, stretched salto forward with 1 turn off.}
        {show with 1 turn.}

\example{(VT) round-off, flic-flac with 0.5 turn on, stretched salto forward with 1.5 turn off.}
        {(VT) round-off, flic-flac with 0.5 turn on, 0.5 turn to piked salto backward off.}
        {show 0.5 turn with spiked salto backward.}

\example{(VT) round-off, flic-flac with 0.5 turn on, stretched salto forward with 1.5 turn off.}
        {(VT) round-off, flic-flac with 1 turn on, piked salto backward off.}
        {show flic-flac with 1 turn and picked salto backward.}

\example{(VT) round-off, flic-flac with 0.5 turn on, stretched salto forward with 0.5 turn off.}
        {(VT) round-off, flic-flac with 0.5 turn on, stretched salto forward with 1.5 turn off.}
        {show with 1.5 turn.}

\example{(VT) round-off, flic-flac with 0.5 turn on, piked salto forward off.}
        {(VT) round-off, flic-flac with 0.5 turn on, stretched salto forward with 1.5 turn off.}
        {show stretched salto forward with 1.5 turn.}

\example{(VT) round-off, flic-flac with 0.5 turn on, piked salto forward off.}
        {(VT) round-off, flic-flac with 1 turn on, piked salto backward off.}
        {show flic-flac with 1 turn and piked salto backward.}

\example{(VT) round-off, flic-flac with 1 turn on, piked salto backward off.}
        {(VT) round-off, flic-flac with 0.5 turn on, piked salto forward off.}
        {show flic-flac with 0.5 turn and piked salto forward.}

\example{(VT) tsukahara stretched with 2 turn.}
        {(VT) tsukahara stretched with 1 turn.}
        {show with 1 turn.}

\example{(VT) tsukahara stretched with 2 turn.}
        {(VT) tsukahara tucked with 1 turn.}
        {show tucked with 1 turn.}

\example{(VT) tsukahara stretched salto.}
        {(VT) tsukahara stretched without salto.}
        {show without salto.}

\example{(FX) switch leap with 0.5 turn.}
        {(BB) switch leap with 0.5 turn.}
        {show on BB.}

\example{(FX) switch leap with 0.5 turn.}
        {(FX) split jump with 0.5 turn.}
        {show a split jump.}

\example{(FX) switch leap with 0.5 turn.}
        {(FX) switch leap.}
        {show a switch leap with no turn.}

\example{(FX) switch leap with 1 turn.}
        {(BB) split leap with 1 turn.}
        {show a split leap on BB.}

\example{(FX) stag jump.}
        {(FX) stag ring jump.}
        {show with ring.}

\example{(FX) tuck hop or jump with 1 turn.}
        {(FX) wolf hop or jump with 1 turn.}
        {show wolf hop.}

\example{(FX) pike jump with 1 turn.}
        {(BB) straddle pike jump with 1 turn.}
        {show straddle pike jump on BB.}

\example{(UB) (swing forward) salto backward stretched.}
        {(UB) (swing backward) double salto forward tucked with 0.5 turn.}
        {show (swing backward) double salto forward tucked with 0.5 turn.}

\example{(UB) (swing forward) double salto backward stretched with 1 turn.}
        {(UB) (swing forward) salto backward stretched with 2 turn.}
        {show salto backward stretched with 2 turn.}



\section{Limitations of Existing Captioning Models}

We present a detailed comparison between the captions generated by existing video captioning models and the structured descriptions curated for our \textit{TF-CoVR} dataset. As \textit{TF-CoVR} is designed around triplets centered on fine-grained temporal actions, it is essential that captioning models capture key elements such as action type, number of turns, and the apparatus involved. Our analysis shows that current models, such as LaVila~\cite{zhao2023learning} and VideoXL~\cite{shu2024video}, often fail to identify these fine-grained details, underscoring their limitations in handling temporally precise and action-specific scenarios.

\paragraph{Caption Generation Template for VideoXL}
To generate technically accurate captions for gymnastics and diving routines, we supply VideoXL with domain-specific prompts tailored to each sport. These prompts incorporate specialized vocabulary and structured syntax to align with official judging terminology. In both sports, subtle variations, such as differences in twist count, body position, or apparatus, convey distinct semantic meaning. To capture this level of granularity, we apply strict formatting constraints and exemplar-based guidance during prompting. While this structured approach helps VideoXL focus on fine-grained action details, the generated captions still exhibit inconsistencies and often fail to capture critical aspects of the routines with sufficient reliability.

\begin{tcolorbox}[title=\textbf{VideoXL Caption Generation Prompt for FineGym}, colback=gray!5, colframe=black!30, coltitle=black, fonttitle=\bfseries, sharp corners, boxrule=0.5pt, arc=3pt]


You are an expert gymnastics judge. 

Your task is to provide a \textbf{strictly formatted, concise technical caption} for the gymnast's  routine. Use \textbf{official gymnastics vocabulary only} (e.g., round-off, flic-flac, salto, tuck, pike, layout).

DO NOT describe emotions, strength, balance, or control.

DO NOT explain what it "shows" or "demonstrates." 

DO NOT use generic verbs like "move", "flip", "spin", "pose", etc.

Include:

- Entry move (e.g., round-off)

- Main move (e.g., double back salto)

- Body position (e.g., tuck, layout, pike)

- Number of twists or somersaults (e.g., 1.5 twists, triple salto)

- Apparatus name if identifiable

Only output a \textbf{single-line caption}, no lists, no bullets, no extra sentences.

Format: 
[Technical move sequence with turns and position]. 

(Apparatus: [FX / VT / BB / UB / Unknown]) 

Examples:

- Round-off, flic-flac, double tuck salto with 1.5 twist. (Apparatus: FX)

- Back handspring to layout salto with full twist. (Apparatus: BB)

- Stretched salto backward with 2.5 twists. (Apparatus: VT)


\end{tcolorbox}

\begin{tcolorbox}[title=\textbf{VideoXL Caption Generation Prompt for FineDiving}, colback=gray!5, colframe=black!30, coltitle=black, fonttitle=\bfseries, sharp corners, boxrule=0.5pt, arc=3pt]


You are an expert \textbf{diving judge}.

Your task is to provide a \textbf{strictly formatted, concise technical caption} for the diver's routine based on official diving terminology. Use terms defined by \textbf{FINA} and standard competition vocabulary.

DO NOT describe emotions, grace, beauty, or control.  
DO NOT narrate or explain what it "shows" or "demonstrates."

DO NOT use vague verbs like "moves", "flips", "spins", or any stylistic language.

Include:

- Takeoff direction (e.g., forward, backward, reverse, inward, armstand)

- Number of somersaults (e.g., 1.5, 2.5, 3.5)

- Number of twists (if any)

- Body position (tuck, pike, layout, free)

- Entry type if clear (e.g., vertical entry, feet-first)

- Platform or springboard (if inferable), e.g., 10m platform, 3m springboard

Only output a \textbf{single-line caption}, no bullets, no extra explanation.

Format:

[Takeoff type], [\# somersaults] somersaults, [\# twists if any] twists, [body position]. 

(Platform: [10m / 3m / Unknown])

Examples:

- Backward takeoff, 2.5 somersaults, tuck. (Platform: 10m)

- Reverse takeoff, 1.5 somersaults, 1 twist, pike. (Platform: 3m)

- Armstand, 2.5 somersaults, layout. (Platform: Unknown)


\end{tcolorbox}

\paragraph{Caption Generation Template for LaViLa}
As an alternative to VideoXL, we also experimented with LaViLa~\cite{zhao2023learning}, a general-purpose multimodal model, to generate captions for both query and target videos. We selected LaViLa based on its prior application in EgoCVR~\cite{hummel2024egocvr}, a task closely related to CoVR. However, the captions produced by LaViLa lacked the fine-grained detail and domain-specific terminology needed to accurately describe gymnastics and diving routines. This gap is illustrated in Table~\ref{tab:finegym_caption_comparison} and Table~\ref{tab:finediving_caption_comparison}, which compare the official label descriptions from \textit{FineGym}~\cite{shao2020finegym} and \textit{FineDiving}~\cite{xu2022finediving} with captions generated by LaViLa and VideoXL.

\vspace{-3em}
\begin{table}[h]
\centering
\caption{Comparison between ground-truth action labels from FineGym and the captions generated by LaViLa and VideoXL. The examples illustrate the inability of both models, particularly LaViLa, to capture fine-grained, domain-specific details such as action type, twist count, and apparatus, which are critical for tasks like \textit{TF-CoVR}.}
\label{tab:finegym_caption_comparison}
\setlength{\tabcolsep}{4pt}
\begin{tabular}{|p{0.31\textwidth}|p{0.31\textwidth}|p{0.31\textwidth}|}
\hline
\parbox[c]{0.31\textwidth}{\centering \textbf{Ground-Truth Label}} & 
\parbox[c]{0.31\textwidth}{\centering \textbf{LaViLa Caption}} & 
\parbox[c]{0.31\textwidth}{\centering \textbf{VideoXL Caption}} \\
\hline

(Vault) round-off, flic-flac with 0.5 turn on, stretched salto forward with 0.5 turn off & \#O A man Y walks around the game & Action: Back Handstand, Turns: 2 \\
\hline
(Vault) round-off, flic-flac on, stretched salto backward with 1 turn off & \#O person X runs on the ground & Action: Flip, Turns: 3 \\
\hline
(Floor Exercise) switch leap with 0.5 turn & \#O The woman A runs towards the woman Y & Action: Flip on the floor, Turns: 3 \\
\hline
(Floor Exercise) switch leap with 1 turn & \#O The man Y jumps down from the wall & Action: Handstand walk with hand release, Turns: 3 \\
\hline
(Floor Exercise) johnson with additional 0.5 turn & \#O The man Y runs towards the man X & Action: Flip, Turns: 0, Action: Dive, Turns: 0 \\
\hline
(Floor Exercise) 2 turn in back attitude, knee of free leg at horizontal throughout turn & \#O The woman B falls to the floor & Action: Twirl, Turns: 0 \\
\hline
(Floor Exercise) 3 turn on one leg, free leg optional below horizontal & \#O The woman Y walks away from the woman X & Action: Flip, Turns: 1 \\
\hline
(Floor Exercise) salto forward tucked & \#O The woman A raises her hands up & Action: Handstand, Turns: 4 \\
\hline
(Floor Exercise) salto forward stretched with 1 twist & \#O The woman X throws the ball with the tennis & Action: Handstand on Rungs, Turns: 15 \\
\hline
(Floor Exercise) salto backward stretched with 3 twist & \#O The man Y throws the slate in his right hand to the ground & Action: Jump from Bar, Turns: 2 \\
\hline
\end{tabular}
\end{table}

\vspace{-4em}
\begin{table}[h]
\centering
\caption{Comparison between ground-truth action labels from FineDiving and captions generated by LaViLa and VideoXL. The examples highlight both models' limitations in capturing critical diving-specific details such as somersault count, twist degree, and entry type. While VideoXL occasionally identifies general action categories, it often fails to reflect the structured semantics required for fine-grained tasks like \textit{TF-CoVR}.}

\label{tab:finediving_caption_comparison}
\setlength{\tabcolsep}{4pt}
\begin{tabular}{|p{0.31\textwidth}|p{0.31\textwidth}|p{0.31\textwidth}|}
\hline
\parbox[c]{0.31\textwidth}{\centering \textbf{Ground-Truth Label}} & 
\parbox[c]{0.31\textwidth}{\centering \textbf{LaViLa Caption}} & 
\parbox[c]{0.31\textwidth}{\centering \textbf{VideoXL Caption}} \\
\hline

Arm.Forward, 2 Soms.Pike, 3.5 Twists & \#O The man X jumps down from the playground slide & Action: Diving, Backflip, Half Turn, T-Walk, Kick flip, Headstand, Handstand, Turns: 3 \\
\hline
Arm.Back, 1.5 Twists, 2 Soms.Pike, Entry & \#O The girl X jumps down from the playhouse & Action: Flip, Turns: 2 \\
\hline
Arm.Back, 2.5 Twists, 2 Soms.Pike, Entry & \#O The man X walks down a stair with the rope in his right hand & Action: Gymnasty Turn, Turns: 4 \\
\hline
Inward, 3.5 Soms.Pike, Entry & \#C C looks at the person in the swimming & Action: Backflip, Turns: 2 \\
\hline
Forward, 3.5 Soms.Pike, Entry & \#C C shakes his right hand & Action: Dive, Turns: 2 \\
\hline
\end{tabular}
\end{table}
\vspace{0em}

Although LaViLa performs well on general video-language benchmarks, it lacks the domain-specific understanding necessary to capture the structured and fine-grained nature of \textit{TF-CoVR} videos. In contrast, targeted prompting with VideoXL produces more consistent and detailed captions, yet it still falls short in accurately identifying the specific actions depicted in \textit{TF-CoVR}.

\section{Experimental Setup}

We evaluate \textit{TF-CoVR} using retrieval-specific metrics, namely mean Average Precision at K (\textit{mAP@K}) for $K \in \{5, 10, 25, 50\}$. All models are trained and evaluated on the \textit{TF-CoVR} dataset using varying video-text encoding strategies and fusion mechanisms.

\paragraph{Video and Text Input Settings.}
We sample 12 uniformly spaced frames from each video and resize them to fit the input dimensions of the pretrained visual backbones. For text input, the modification texts are tokenized using the tokenizer corresponding to each text encoder (e.g., CLIP or BLIP) and passed to the model without truncation whenever possible.

\paragraph{Text Encoder Evaluation.}
To evaluate the impact of different text encoders on the \textit{TF-CoVR-Base} model, we conducted experiments using two popular pretrained vision-language models: CLIP and BLIP. Both models were used to encode the \textit{modification text} inputs, while the visual backbone and fusion mechanism were held constant (MLP-based fusion with 12-frame video inputs). As shown in Table~\ref{tab:wv_benchmark}, BLIP consistently outperforms CLIP across all \textit{mAP@K} metrics, suggesting a stronger ability to capture the semantic nuances of the modification texts. Each experiment was repeated five times, and we report the mean and standard deviation to ensure robustness.

\begin{table}[h]
\centering
\caption{Evaluation of \textit{TF-CoVR-Base} fine-tuned on \textit{TF-CoVR} with different text encoders using mAP@K for $K \in \{5, 10, 25, 50\}$. We ran each experiment five times and report mean and standard deviation in the following table}
\vspace{0.5em}
\label{tab:wv_benchmark}
\setlength{\tabcolsep}{3pt}
\resizebox{\textwidth}{!}{%
\begin{tabular}{
    cc     
    c      
    c      
    c      
    c c    
    c c c c
}
\toprule
\multicolumn{2}{c}{\textbf{Modalities}} 
  & \textbf{Model}
  & \textbf{Text}
  & \textbf{Fusion}
  & \#\textbf{Query}
  & \#\textbf{Target}
  & \multicolumn{4}{c}{\textbf{mAP@K ($\uparrow$)}} \\
\cmidrule(lr){1-2} \cmidrule(lr){8-11}
\textbf{Video} & \textbf{Text} 
  &  & \textbf{Encoder} &  & \textbf{Frames} & \textbf{Frames}
  & 5 & 10 & 25 & 50 \\

\midrule

\cmark & \cmark
  & TF-CoVR-Base & CLIP & MLP & 12 & 12
  & 18.30 $\pm$ 0.35 & 20.59 $\pm$ 0.30 & 22.89 $\pm$ 0.27 & 23.64 $\pm$ 0.27 \\ 

\cmark & \cmark
  & TF-CoVR-Base & BLIP & MLP & 12 & 12
  & 20.62 $\pm$ 0.25 & 23.17 $\pm$ 0.34 & 25.17 $\pm$ 0.28 & 25.88 $\pm$ 0.25 \\ 
\bottomrule
\end{tabular}
}

\end{table}

\paragraph{Fusion Module.}
We use a lightweight multi-layer perceptron (MLP) with two hidden layers and ReLU activation to combine visual and textual features, enabling efficient multimodal fusion while preserving architectural simplicity.

\paragraph{Training and Evaluation Protocols.}
We fine-tune each model using the AdamW optimizer with a learning rate of $1 \times 10^{-4}$ and a batch size of 512. Each model is trained for 100 epochs. All configurations are evaluated across five random seeds to ensure statistical reliability.

\paragraph{Hardware Configuration and Training Time.}
All experiments were conducted on four NVIDIA A100 GPUs, each with 80 GB of memory. Stage 1 pretraining, performed on two datasets using a single A100 GPU, takes approximately four days, while Stage 2 fine-tuning completes in about six hours. 


\section{TF-CoVR Visualization}
\begin{figure}[t] 
    \centering 
    \includegraphics[width=0.95\textwidth]{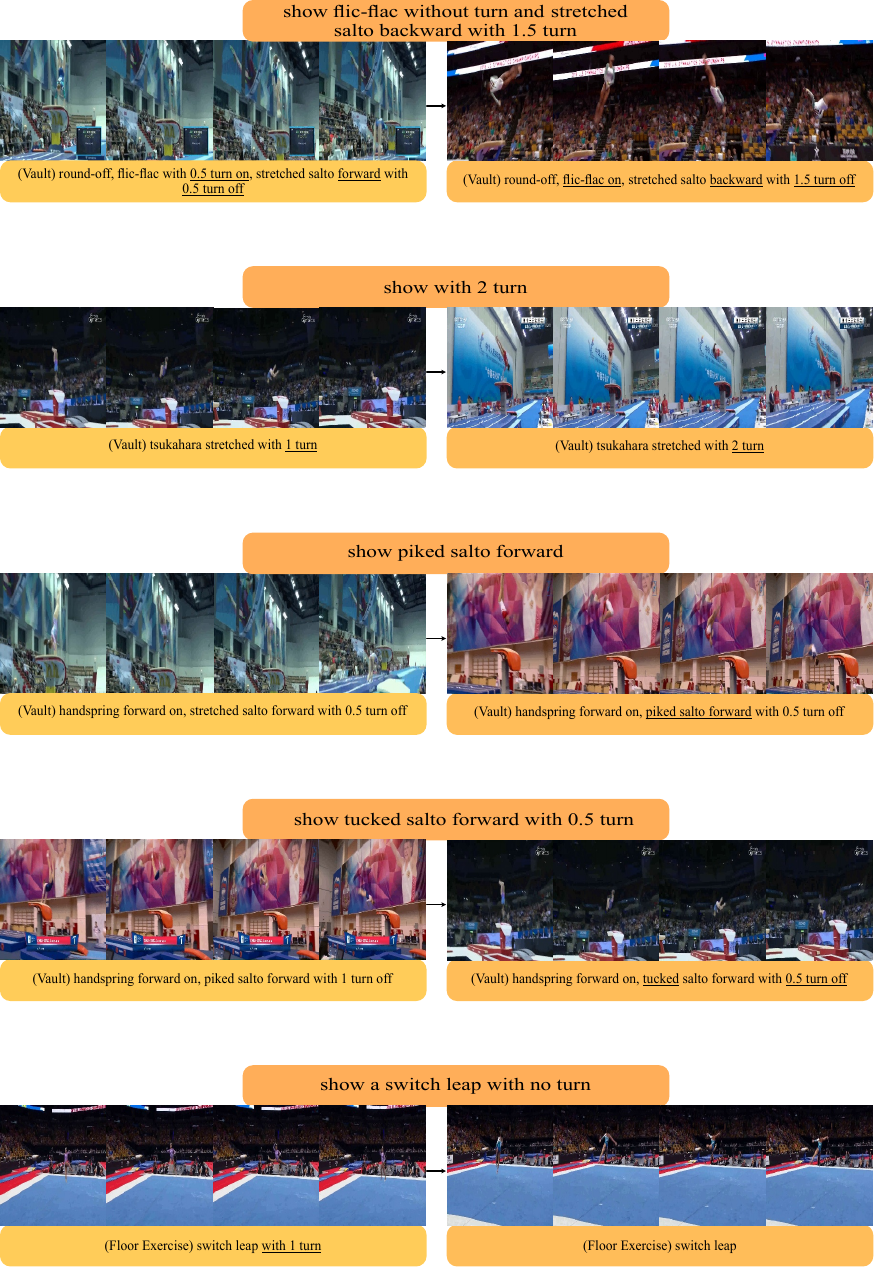} 
    \caption{Qualitative examples from \textit{TF-CoVR} showcasing motion-centric transformations for fine-grained temporal action retrieval. The examples span diverse gymnastic events such as \textit{vaults} and \textit{floor exercises}, where subtle differences in execution such as changing from a \textit{stretched} to a \textit{tucked salto}, increasing the number of turns from \textit{one} to \textit{two}, or removing rotation in a \textit{switch leap} define the compositional shift. The captions explicitly highlight these movement attributes, enabling precise instruction-based retrieval grounded in temporal dynamics rather than visual appearance or scene context. This focus on action semantics and minimal visual distraction distinguishes \textit{TF-CoVR} from prior CoVR datasets.}
    \label{fig:supp_1} 
\end{figure}

\begin{figure}[t] 
    \centering 
    \includegraphics[width=0.95\textwidth]{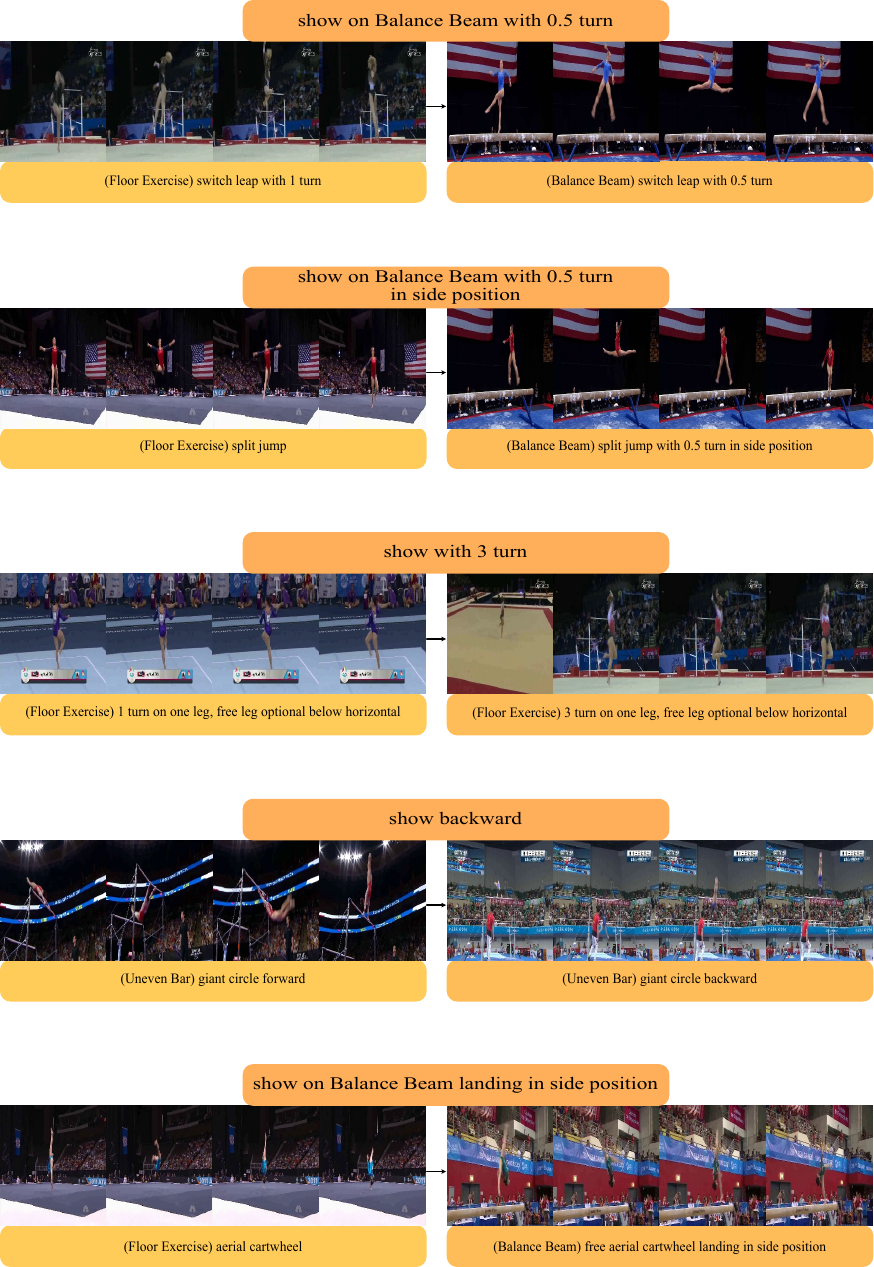} 
    \caption{Additional examples from \textit{TF-CoVR} demonstrating temporally grounded modifications across multiple apparatuses. Each triplet reflects precise motion-based transformations driven by modification instructions, such as “\textit{show with 3 turn}”, “\textit{show on Balance Beam with 0.5 turn in side position}”, or “\textit{show backward}”.}
    \label{fig:supp_2} 
\end{figure}

\begin{figure}[t] 
    \centering 
    \includegraphics[width=0.95\textwidth]{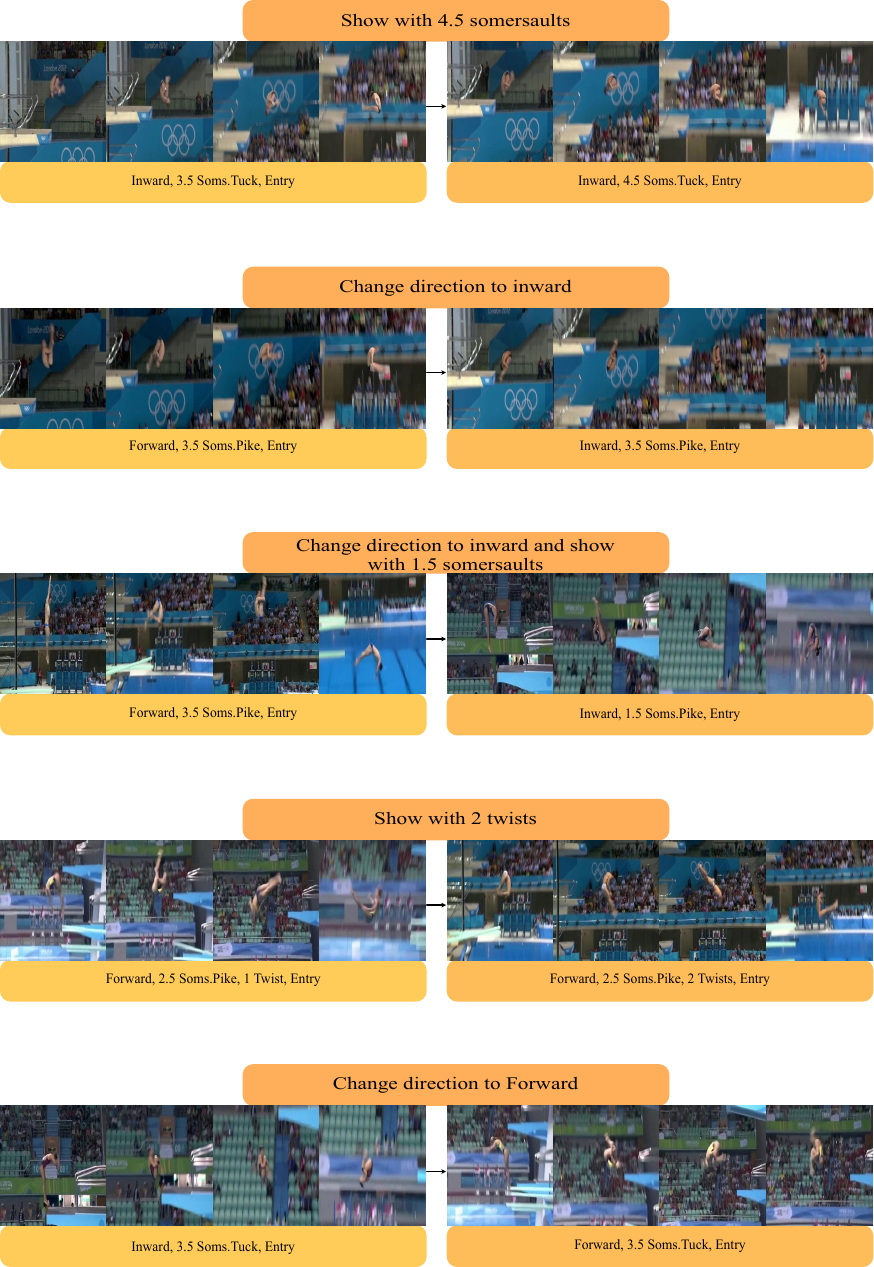} 
    \caption{\textit{TF-CoVR} triplets from diving events demonstrating precise compositional modifications based on somersault count, twist count, and direction. Examples include transformations such as “\textit{Show with 4.5 somersaults},” “\textit{Change direction to inward}”, “\textit{Change direction to inward and show with 1.5 somersaults}”, “\textit{Show with 2 twists}”, and “\textit{Change direction to forward}”. Each caption specifies critical motion semantics like entry type, direction (\textit{forward} or \textit{inward}), somersault type (\textit{Tuck} or \textit{Pike}), and twist count, enabling controlled retrieval grounded in temporally fine-grained action variations.}
    \label{fig:supp_3} 
\end{figure}

\textit{TF-CoVR} (Figure~\ref{fig:supp_1}) offers a clear, structured visualization of the Composed Video Retrieval (CoVR) task, specifically designed for fine-grained temporal understanding. Unlike prior CoVR benchmarks such as WebVid CoVR~\cite{ventura2024covr} and EgoCVR~\cite{hummel2024egocvr}, which often rely on broad scene-level changes or object variations, \textit{TF-CoVR} centers on subtle, motion-centric transformations. These include variations in the number of turns, transitions between salto types (e.g., \textit{tucked}, \textit{piked}, or \textit{stretched}), and the inclusion or omission of rotational components in gymnastic leaps.

Each row in the figure illustrates a triplet: the left column displays the \textit{query video}, the right shows the corresponding \textit{target video}, and the center presents the \textit{modification text} describing the transformation required to reach the target. \textit{TF-CoVR} emphasizes action-specific, apparatus-consistent changes, where even subtle variations in movement or rotation denote semantically distinct actions. By controlling for background and scene context, the figure isolates fine-grained motion differences as the primary signal for retrieval. This makes \textit{TF-CoVR} a strong benchmark for assessing whether models can accurately retrieve videos based on instruction-driven, temporally grounded modifications. Additional visualizations of \textit{TF-CoVR} are provided in Figures~\ref{fig:supp_2} and~\ref{fig:supp_3}.

\section{Institutional Review Board (IRB) Approval}

\textit{TF-CoVR} uses publicly available videos from the \textit{FineGym} and \textit{FineDiving} datasets. Access to these videos is subject to the licensing terms specified by the respective dataset providers. To support reproducibility, we released the video and text embeddings generated during our experiments.

\begin{table}[h]
    \centering
    \resizebox{\textwidth}{!}{%
    \begin{tabular}{|p{0.05\textwidth}|p{0.28\textwidth}|p{0.05\textwidth}|p{0.28\textwidth}|p{0.05\textwidth}|p{0.28\textwidth}|}
    \hline
        Label 1 & Caption 1 & Label 2 & Caption 2 & Label 3 & Caption 3 \\ \hline
        0 & (Vault) round-off, flic-flac with 0.5 turn on, stretched salto forward with 1.5 turn off & 1 & (Vault) round-off, flic-flac with 0.5 turn on, stretched salto forward with 0.5 turn off & 2 & (Vault) round-off, flic-flac with 0.5 turn on, stretched salto forward with 1 turn off \\ \hline
        3 & (Vault) round-off, flic-flac with 0.5 turn on, stretched salto forward with 2 turn off & 4 & (Vault) round-off, flic-flac with 0.5 turn on, 0.5 turn to piked salto backward off & 5 & (Vault) round-off, flic-flac with 0.5 turn on, piked salto forward with 0.5 turn off \\ \hline
        6 & (Vault) round-off, flic-flac with 0.5 turn on, piked salto forward off & 7 & (Vault) round-off, flic-flac with 0.5 turn on, tucked salto forward with 0.5 turn off & 8 & (Vault) round-off, flic-flac with 1 turn on, stretched salto backward with 1 turn off \\ \hline
        9 & (Vault) round-off, flic-flac with 1 turn on, piked salto backward off & 10 & (Vault) round-off, flic-flac on, stretched salto backward with 2 turn off & 11 & (Vault) round-off, flic-flac on, stretched salto backward with 1 turn off \\ \hline
        12 & (Vault) round-off, flic-flac on, stretched salto backward with 1.5 turn off & 13 & (Vault) round-off, flic-flac on, stretched salto backward with 0.5 turn off & 14 & (Vault) round-off, flic-flac on, stretched salto backward with 2.5 turn off \\ \hline
        15 & (Vault) round-off, flic-flac on, stretched salto backward off & 16 & (Vault) round-off, flic-flac on, piked salto backward off & 17 & (Vault) round-off, flic-flac on, tucked salto backward off \\ \hline
        18 & (Vault) tsukahara stretched with 2 turn & 19 & (Vault) tsukahara stretched with 1 turn & 20 & (Vault) tsukahara stretched with 1.5 turn \\ \hline
        21 & (Vault) tsukahara stretched with 0.5 turn & 22 & (Vault) tsukahara stretched salto & 23 & (Vault) tsukahara stretched without salto \\ \hline
        28 & (Vault) tsukahara tucked with 1 turn & 28 & (Vault) handspring forward on, stretched salto forward with 1.5 turn off & 28 & (Vault) handspring forward on, stretched salto forward with 0.5 turn off \\ \hline
        29 & (Vault) handspring forward on, stretched salto forward with 1 turn off & 28 & (Vault) handspring forward on, piked salto forward with 0.5 turn off & 31 & (Vault) handspring forward on, piked salto forward with 1 turn off \\ \hline
        32 & (Vault) handspring forward on, piked salto forward off & 33 & (Vault) handspring forward on, tucked salto forward with 0.5 turn off & 34 & (Vault) handspring forward on, tucked salto forward with 1 turn off \\ \hline
        35 & (Vault) handspring forward on, tucked double salto forward off & 36 & (Vault) handspring forward on, tucked salto forward off & 37 & (Vault) handspring forward on, 1.5 turn off \\ \hline
        38 & (Vault) handspring forward on, 1 turn off & 40 & (Floor Exercise) switch leap with 0.5 turn & 41 & (Floor Exercise) switch leap with 1 turn \\ \hline
        42 & (Floor Exercise) split leap with 0.5 turn & 43 & (Floor Exercise) split leap with 1 turn & 44 & (Floor Exercise) split leap with 1.5 turn or more \\ \hline
        45 & (Floor Exercise) switch leap & 46 & (Floor Exercise) split leap forward & 47 & (Floor Exercise) split jump with 1 turn \\ \hline
        48 & (Floor Exercise) split jump with 0.5 turn & 49 & (Floor Exercise) split jump with 1.5 turn & 51 & (Floor Exercise) split jump \\ \hline
        52 & (Floor Exercise) johnson with additional 0.5 turn & 53 & (Floor Exercise) johnson & 54 & (Floor Exercise) straddle pike or side split jump with 1 turn \\ \hline
        55 & (Floor Exercise) straddle pike or side split jump with 0.5 turn & 56 & (Floor Exercise) straddle pike jump or side split jump & 57 & (Floor Exercise) stag ring jump \\ \hline
        58 & (Floor Exercise) switch leap to ring position with 1 turn & 59 & (Floor Exercise) switch leap to ring position & 60 & (Floor Exercise) split leap with 1 turn or more to ring position \\ \hline
        61 & (Floor Exercise) split ring leap & 62 & (Floor Exercise) ring jump & 63 & (Floor Exercise) split jump with 1 turn or more to ring position \\ \hline
        65 & (Floor Exercise) stag jump & 66 & (Floor Exercise) tuck hop or jump with 1 turn & 67 & (Floor Exercise) tuck hop or jump with 2 turn \\ \hline
        68 & (Floor Exercise) stretched hop or jump with 1 turn & 69 & (Floor Exercise) pike jump with 1 turn & 70 & (Floor Exercise) sheep jump \\ \hline
        71 & (Floor Exercise) wolf hop or jump with 1 turn & 73 & (Floor Exercise) wolf hop or jump & 76 & (Floor Exercise) cat leap \\ \hline
        77 & (Floor Exercise) hop with 0.5 turn free leg extended above horizontal throughout & 78 & (Floor Exercise) hop with 1 turn free leg extended above horizontal throughout & 81 & (Floor Exercise) 3 turn with free leg held upward in 180 split position throughout turn \\ \hline
        
    \end{tabular}
    }
\end{table}

\begin{table}[h]
    \centering
    \resizebox{\textwidth}{!}{%
    \begin{tabular}{|p{0.05\textwidth}|p{0.28\textwidth}|p{0.05\textwidth}|p{0.28\textwidth}|p{0.05\textwidth}|p{0.28\textwidth}|}
    \hline
        Label 1 & Caption 1 & Label 2 & Caption 2 & Label 3 & Caption 3 \\ \hline
        82 & (Floor Exercise) 2 turn with free leg held upward in 180 split position throughout turn & 83 & (Floor Exercise) 1 turn with free leg held upward in 180 split position throughout turn & 84 & (Floor Exercise) 3 turn in tuck stand on one leg, free leg straight throughout turn \\ \hline
        85 & (Floor Exercise) 2 turn in tuck stand on one leg, free leg straight throughout turn & 86 & (Floor Exercise) 1 turn in tuck stand on one leg, free leg optional & 88 & (Floor Exercise) 2 turn in back attitude, knee of free leg at horizontal throughout turn \\ \hline
        89 & (Floor Exercise) 1 turn in back attitude, knee of free leg at horizontal throughout turn & 90 & (Floor Exercise) 4 turn on one leg, free leg optional below horizontal & 91 & (Floor Exercise) 3 turn on one leg, free leg optional below horizontal \\ \hline
        92 & (Floor Exercise) 2 turn on one leg, free leg optional below horizontal & 93 & (Floor Exercise) 1 turn on one leg, free leg optional below horizontal & 94 & (Floor Exercise) 2 turn or more with heel of free leg forward at horizontal throughout turn \\ \hline
        95 & (Floor Exercise) 1 turn with heel of free leg forward at horizontal throughout turn & 97 & (Floor Exercise) aerial cartwheel & 98 & (Floor Exercise) arabian double salto tucked \\ \hline
        99 & (Floor Exercise) double salto forward tucked with 0.5 twist & 100 & (Floor Exercise) double salto forward tucked & 101 & (Floor Exercise) salto forward tucked \\ \hline
        102 & (Floor Exercise) arabian double salto piked & 105 & (Floor Exercise) double salto forward piked & 104 & (Floor Exercise) salto forward piked \\ \hline
        105 & (Floor Exercise) aerial walkover forward & 106 & (Floor Exercise) salto forward stretched with 2 twist & 107 & (Floor Exercise) salto forward stretched with 1 twist \\ \hline
        108 & (Floor Exercise) salto forward stretched with 1.5 twist & 109 & (Floor Exercise) salto forward stretched with 0.5 twist & 110 & (Floor Exercise) salto forward stretched, feet land successively \\ \hline
        111 & (Floor Exercise) salto forward stretched, feet land together & 112 & (Floor Exercise) double salto backward stretched with 2 twist & 113 & (Floor Exercise) double salto backward stretched with 1 twist \\ \hline
        114 & (Floor Exercise) double salto backward stretched with 0.5 twist & 115 & (Floor Exercise) double salto backward stretched & 116 & (Floor Exercise) salto backward stretched with 3 twist \\ \hline
        117 & (Floor Exercise) salto backward stretched with 2 twist & 118 & (Floor Exercise) salto backward stretched with 1 twist & 119 & (Floor Exercise) salto backward stretched \\ \hline
        120 & (Floor Exercise) salto backward stretched with 3.5 twist & 121 & (Floor Exercise) salto backward stretched with 2.5 twist & 122 & (Floor Exercise) salto backward stretched with 1.5 twist \\ \hline
        123 & (Floor Exercise) salto backward stretched with 0.5 twist & 124 & (Floor Exercise) double salto backward tucked with 2 twist & 128 & (Floor Exercise) double salto backward tucked with 1 twist \\ \hline
        126 & (Floor Exercise) double salto backward tucked & 128 & (Floor Exercise) salto backward tucked & 128 & (Floor Exercise) double salto backward piked with 1 twist \\ \hline
        129 & (Floor Exercise) double salto backward piked & 133 & (Balance Beam) split jump with 0.5 turn in side position & 134 & (Balance Beam) split jump with 0.5 turn \\ \hline
        135 & (Balance Beam) split jump with 1 turn & 136 & (Balance Beam) split jump & 137 & (Balance Beam) straddle pike jump with 0.5 turn in side position \\ \hline
        138 & (Balance Beam) straddle pike jump with 0.5 turn & 139 & (Balance Beam) straddle pike jump with 1 turn & 140 & (Balance Beam) straddle pike jump or side split jump in side position \\ \hline
        141 & (Balance Beam) straddle pike jump or side split jump & 142 & (Balance Beam) stag-ring jump & 143 & (Balance Beam) ring jump \\ \hline
        144 & (Balance Beam) split ring jump & 145 & (Balance Beam) switch leap with 0.5 turn & 146 & (Balance Beam) switch leap with 1 turn \\ \hline
        147 & (Balance Beam) split leap with 1 turn & 148 & (Balance Beam) switch leap & 150 & (Balance Beam) split leap forward \\ \hline
        151 & (Balance Beam) johnson with additional 0.5 turn & 152 & (Balance Beam) johnson & 153 & (Balance Beam) switch leap to ring position \\ \hline
        154 & (Balance Beam) split ring leap & 155 & (Balance Beam) tuck hop or jump with 1 turn & 156 & (Balance Beam) tuck hop or jump with 0.5 turn \\ \hline
        158 & (Balance Beam) stretched jump/hop with 1 turn & 159 & (Balance Beam) sheep jump & 160 & (Balance Beam) wolf hop or jump with 1 turn \\ \hline
        161 & (Balance Beam) wolf hop or jump with 0.5 turn & 162 & (Balance Beam) wolf hop or jump & 163 & (Balance Beam) cat leap \\ \hline
        
    \end{tabular}
    }
\end{table}

\begin{table}[h]
    \centering
    \resizebox{\textwidth}{!}{%
    \begin{tabular}{|p{0.05\textwidth}|p{0.28\textwidth}|p{0.05\textwidth}|p{0.28\textwidth}|p{0.05\textwidth}|p{0.28\textwidth}|}
    \hline
        Label 1 & Caption 1 & Label 2 & Caption 2 & Label 3 & Caption 3 \\ \hline
        165 & (Balance Beam) 1.5 turn with free leg held upward in 180 split position throughout turn & 166 & (Balance Beam) 1 turn with free leg held upward in 180 split position throughout turn & 167 & (Balance Beam) 1.5 turn with heel of free leg forward at horizontal throughout turn \\ \hline
        168 & (Balance Beam) 2 turn with heel of free leg forward at horizontal throughout turn & 169 & (Balance Beam) 1 turn with heel of free leg forward at horizontal throughout turn & 170 & (Balance Beam) 2 turn on one leg, free leg optional below horizontal \\ \hline
        171 & (Balance Beam) 1.5 turn on one leg, free leg optional below horizontal & 172 & (Balance Beam) 1 turn on one leg, free leg optional below horizontal & 173 & (Balance Beam) 1 turn on one leg, thigh of free leg at horizontal, backward upward throughout turn \\ \hline
        174 & (Balance Beam) 2.5 turn in tuck stand on one leg, free leg optional & 175 & (Balance Beam) 1.5 turn in tuck stand on one leg, free leg optional & 176 & (Balance Beam) 3 turn in tuck stand on one leg, free leg optional \\ \hline
        177 & (Balance Beam) 2 turn in tuck stand on one leg, free leg optional & 178 & (Balance Beam) 1 turn in tuck stand on one leg, free leg optional & 179 & (Balance Beam) jump forward with 0.5 twist and salto backward tucked \\ \hline
        180 & (Balance Beam) salto backward tucked with 1 twist & 181 & (Balance Beam) salto backward tucked & 182 & (Balance Beam) salto backward piked \\ \hline
        183 & (Balance Beam) gainer salto backward stretched-step out (feet land successively) & 184 & (Balance Beam) salto backward stretched-step out (feet land successively) & 185 & (Balance Beam) salto backward stretched with 1 twist \\ \hline
        186 & (Balance Beam) salto backward stretched with legs together & 187 & (Balance Beam) salto sideward tucked with 0.5 turn, take off from one leg to side stand & 188 & (Balance Beam) salto sideward tucked, take off from one leg to side stand \\ \hline
        189 & (Balance Beam) free aerial cartwheel landing in side position & 191 & (Balance Beam) free aerial cartwheel landing in cross position & 192 & (Balance Beam) arabian salto tucked \\ \hline
        193 & (Balance Beam) salto forward tucked to cross stand & 194 & (Balance Beam) salto forward piked to cross stand & 195 & (Balance Beam) salto forward tucked (take-off from one leg to stand on one or two feet) \\ \hline
        196 & (Balance Beam) free aerial walkover forward, landing on one or both feet & 197 & (Balance Beam) flic-flac with 1 twist, swing down to cross straddle sit & 198 & (Balance Beam) flic-flac, swing down to cross straddle sit \\ \hline
        207 & (Balance Beam) arabian double salto forward tucked & 208 & (Balance Beam) salto forward tucked with 1 twist & 209 & (Balance Beam) salto forward tucked \\ \hline
        210 & (Balance Beam) salto forward piked & 211 & (Balance Beam) salto forward stretched with 1.5 twist & 212 & (Balance Beam) salto forward stretched with 1 twist \\ \hline
        213 & (Balance Beam) salto forward stretched & 214 & (Balance Beam) double salto backward tucked with 1 twist & 215 & (Balance Beam) double salto backward tucked \\ \hline
        216 & (Balance Beam) salto backward tucked with 1 twist & 217 & (Balance Beam) salto backward tucked & 218 & (Balance Beam) salto backward tucked with 1.5 twist \\ \hline
        219 & (Balance Beam) double salto backward piked & 220 & (Balance Beam) salto backward stretched with 3 twist & 221 & (Balance Beam) salto backward stretched with 2 twist \\ \hline
        222 & (Balance Beam) salto backward stretched with 1 twist & 223 & (Balance Beam) salto backward stretched & 224 & (Balance Beam) salto backward stretched with 2.5 twist \\ \hline
        228 & (Balance Beam) salto backward stretched with 1.5 twist & 226 & (Balance Beam) salto backward stretched with 0.5 twist & 228 & (Balance Beam) gainer salto backward stretched with 1 twist to side of beam \\ \hline
        228 & (Balance Beam) gainer salto tucked at end of beam & 229 & (Balance Beam) gainer salto piked at end of beam & 228 & (Balance Beam) gainer salto stretched with 1 twist at end of beam \\ \hline
        231 & (Balance Beam) gainer salto stretched with legs together at end of the beam & 232 & (Uneven Bar) pike sole circle backward with 1.5 turn to handstand & 233 & (Uneven Bar) pike sole circle backward with 1 turn to handstand \\ \hline
        234 & (Uneven Bar) pike sole circle backward with 0.5 turn to handstand & 235 & (Uneven Bar) pike sole circle backward to handstand & 236 & (Uneven Bar) pike sole circle forward with 0.5 turn to handstand \\ \hline
        237 & (Uneven Bar) giant circle backward with 1.5 turn to handstand & 238 & (Uneven Bar) giant circle backward with hop 1 turn to handstand & 239 & (Uneven Bar) giant circle backward with 1 turn to handstand \\ \hline
        240 & (Uneven Bar) giant circle backward with 0.5 turn to handstand & 241 & (Uneven Bar) giant circle backward & 242 & (Uneven Bar) giant circle forward with 1 turn on one arm before handstand phase \\ \hline    
    \end{tabular}
    }
\end{table}

\begin{table}[h]
    \centering
    \resizebox{\textwidth}{!}{%
    \begin{tabular}{|p{0.05\textwidth}|p{0.28\textwidth}|p{0.05\textwidth}|p{0.28\textwidth}|p{0.05\textwidth}|p{0.28\textwidth}|}
    \hline
        Label 1 & Caption 1 & Label 2 & Caption 2 & Label 3 & Caption 3 \\ \hline
        243 & (Uneven Bar) giant circle forward with 1 turn to handstand & 244 & (Uneven Bar) giant circle forward with 1.5 turn to handstand & 245 & (Uneven Bar) giant circle forward with 0.5 turn to handstand \\ \hline
        246 & (Uneven Bar) giant circle forward & 247 & (Uneven Bar) clear hip circle backward with 1 turn to handstand & 248 & (Uneven Bar) clear hip circle backward with 0.5 turn to handstand \\ \hline
        249 & (Uneven Bar) clear hip circle backward to handstand & 280 & (Uneven Bar) clear hip circle forward with 0.5 turn to handstand & 281 & (Uneven Bar) clear hip circle forward to handstand \\ \hline
        282 & (Uneven Bar) clear pike circle backward with 1 turn to handstand & 285 & (Uneven Bar) clear pike circle backward with 0.5 turn to handstand & 284 & (Uneven Bar) clear pike circle backward to handstand \\ \hline
        285 & (Uneven Bar) clear pike circle forward to handstand & 286 & (Uneven Bar) stalder backward with 1 turn to handstand & 287 & (Uneven Bar) stalder backward with 0.5 turn to handstand \\ \hline
        288 & (Uneven Bar) stalder backward to handstand & 289 & (Uneven Bar) stalder forward with 0.5 turn to handstand & 260 & (Uneven Bar) stalder forward to handstand \\ \hline
        262 & (Uneven Bar) counter straddle over high bar with 0.5 turn to hang & 263 & (Uneven Bar) counter straddle over high bar to hang & 264 & (Uneven Bar) counter piked over high bar to hang \\ \hline
        266 & (Uneven Bar) (swing backward or front support) salto forward straddled to hang on high bar & 267 & (Uneven Bar) (swing backward) salto forward piked to hang on high bar & 268 & (Uneven Bar) (swing forward or hip circle backward) salto backward with 0.5 turn piked to hang on high bar \\ \hline
        269 & (Uneven Bar) (swing backward) salto forward stretched to hang on high bar & 280 & (Uneven Bar) (swing forward) salto backward stretched with 0.5 turn to hang on high bar & 281 & (Uneven Bar) transition flight from high bar to low bar \\ \hline
        282 & (Uneven Bar) transition flight from low bar to high bar & 285 & (Uneven Bar) (swing forward) double salto backward tucked with 1.5 turn & 284 & (Uneven Bar) (swing forward) salto with 0.5 turn into salto forward tucked \\ \hline
        285 & (Uneven Bar) (swing forward) double salto backward tucked with 2 turn & 286 & (Uneven Bar) (swing forward) double salto backward tucked with 1 turn & 287 & (Uneven Bar) (swing forward) double salto backward tucked \\ \hline
        288 & (Uneven Bar) (swing backward) double salto forward tucked & 289 & (Uneven Bar) (swing backward) salto forward with 0.5 turn & 280 & (Uneven Bar) (swing backward) double salto forward tucked with 0.5 turn \\ \hline
        281 & (Uneven Bar) (under-swing or clear under-swing) salto forward tucked with 0.5 turn & 282 & (Uneven Bar) (swing forward) double salto backward piked & 283 & (Uneven Bar) (swing forward) double salto backward stretched with 2 turn \\ \hline
        284 & (Uneven Bar) (swing forward) double salto backward stretched with 1 turn & 285 & (Uneven Bar) (swing forward) double salto backward stretched & 286 & (Uneven Bar) (swing forward) salto backward stretched with 2 turn \\ \hline
        287 & (Uneven Bar) (swing forward) salto backward stretched & 407c & Inward, 3.5 Soms.Tuck, Entry & 5285b & Back, 1.5 Twists, 2.5 Soms.Pike, Entry \\ \hline
        107b & Forward, 3.5 Soms.Pike, Entry & 6245d & Arm.Back, 2.5 Twists, 2 Soms.Pike, Entry & 207c & Back, 3.5 Soms.Tuck, Entry \\ \hline
        5152b & Forward, 2.5 Soms.Pike, 1 Twist, Entry & 5285b & Back, 2.5 Twists, 2.5 Soms.Pike, Entry & 6243d & Arm.Back, 1.5 Twists, 2 Soms.Pike, Entry \\ \hline
        109c & Forward, 4.5 Soms.Tuck, Entry & 626c & Arm.Back, 3 Soms.Tuck, Entry & 287c & Reverse, 3.5 Soms.Tuck, Entry \\ \hline
        207b & Back, 3.5 Soms.Pike, Entry & 5156b & Forward, 2.5 Soms.Pike, 3 Twists, Entry & 407b & Inward, 3.5 Soms.Pike, Entry \\ \hline
        409c & Inward, 4.5 Soms.Tuck, Entry & 6142d & Arm.Forward, 1 Twist, 2 Soms.Pike, 3.5 Twists & 285c & Reverse, 2.5 Soms.Tuck, Entry \\ \hline
        405b & Inward, 2.5 Soms.Pike, Entry & 205b & Back, 2.5 Soms.Pike, Entry & 5235d & Back, 2.5 Twists, 1.5 Soms.Pike, Entry \\ \hline
        612b & Arm.Forward, 2 Soms.Pike, 3.5 Twists & 105b & Forward, 1.5 Soms.Pike, Entry & 405b & Inward, 1.5 Soms.Pike, Entry \\ \hline
        101b & Forward, 0.5 Som.Pike, Entry & 5331d & Reverse, 0.5 Twist, 1.5 Soms.Pike, Entry & 5132d & Forward, 1.5 Soms.Pike, 1 Twist, Entry \\ \hline
        614b & Arm.Forward, 2 Soms.Pike, Entry & 5231d & Back, 0.5 Twist, 1.5 Soms.Pike, Entry & 5154b & Forward, 2.5 Soms.Pike, 2 Twists, Entry \\ \hline        
    \end{tabular}
    }
\end{table}

\begin{table}[ht]
    \centering
    \resizebox{\textwidth}{!}{%
    \begin{tabular}{|p{0.05\textwidth}|p{0.28\textwidth}|p{0.05\textwidth}|p{0.28\textwidth}|p{0.05\textwidth}|p{0.28\textwidth}|}
    \hline
        Label 1 & Caption 1 & Label 2 & Caption 2 & Label 3 & Caption 3 \\ \hline
        5281b & Back, 1.5 Twists, 2.5 Soms.Pike, Entry & 107c & Forward, 3.5 Soms.Tuck, Entry & 105b & Forward, 2.5 Soms.Pike, Entry \\ \hline
        6241b & Forward, 0.5 Twist, 2 Soms.Pike, Entry & 5237d & Back, 3.5 Twists, 1.5 Soms.Pike, Entry & 5353b & Reverse, 1.5 Twists, 2.5 Soms.Pike, Entry \\ \hline
        5337d & Reverse, 3.5 Twists, 1.5 Soms.Pike, Entry & 5355b & Reverse, 2.5 Twists, 2.5 Soms.Pike, Entry & 405c & Inward, 2.5 Soms.Tuck, Entry \\ \hline
        5335d & Reverse, 2.5 Twists, 1.5 Soms.Pike, Entry & 5172b & Forward, 3.5 Soms.Pike, 1 Twist, Entry & 636c & Arm.Reverse, 3 Soms.Tuck, Entry \\ \hline
        205c & Back, 2.5 Soms.Tuck, Entry & 626b & Arm.Back, 3 Soms.Pike, Entry & 401b & Inward, 0.5 Som.Pike, Entry \\ \hline
        5233d & Back, 1.5 Twists, 1.5 Soms.Pike, Entry & 109b & Forward, 4.5 Soms.Pike, Entry & 285c & Reverse, 1.5 Soms.Tuck, Entry \\ \hline
    \end{tabular}
    }
\end{table}

\end{document}